\pdfoutput=1
\documentclass[nohyperref]{article}

\usepackage{microtype}
\usepackage{graphicx}
\usepackage{subfigure}
\usepackage{booktabs} %
\usepackage{breqn}
\usepackage{multirow}
\usepackage{booktabs, caption, makecell}

\usepackage{threeparttable}
\usepackage{xcolor}
\usepackage{hyperref}
\usepackage{color, colortbl}
\definecolor{Gray}{gray}{0.9}
\newcommand{\ccolor}{\cellcolor{Gray}}

\usepackage[accepted]{icml2022}

\usepackage{amsmath}
\usepackage{amssymb}
\usepackage{mathtools}
\usepackage{amsthm}

\usepackage[capitalize,noabbrev]{cleveref}

\newcommand{\Alg}{BCL}

\theoremstyle{plain}

\theoremstyle{definition}

\theoremstyle{remark}

\usepackage[textsize=tiny]{todonotes}

\allowdisplaybreaks
\icmltitlerunning{Robust Deep Reinforcement Learning through Bootstrapped Opportunistic Curriculum}

\begin{document}

\twocolumn[
\icmltitle{Robust Deep Reinforcement Learning through\\ Bootstrapped Opportunistic Curriculum}

\icmlsetsymbol{equal}{*}

\begin{icmlauthorlist}
\icmlauthor{Junlin Wu}{yyy}
\icmlauthor{Yevgeniy Vorobeychik}{yyy}
\end{icmlauthorlist}

\icmlaffiliation{yyy}{Department of Computer Science and Engineering, Washington University in St. Louis, St. Louis, MO, USA}

\icmlcorrespondingauthor{Junlin Wu}{junlin.wu@wustl.edu}
\icmlcorrespondingauthor{Yevgeniy Vorobeychik}{yvorobeychik@wustl.edu}

\icmlkeywords{Adversarial Machine Learning, Curriculum Learning, Reinforcement Learning, Robustness}

\vskip 0.3in
]

\printAffiliationsAndNotice{}  %
\begin{abstract}
Despite considerable advances in deep reinforcement learning, it has been shown to be highly vulnerable to adversarial perturbations to state observations.
Recent efforts that have attempted to improve adversarial robustness of reinforcement learning can nevertheless tolerate only very small perturbations, and remain fragile as perturbation size increases.
We propose \emph{Bootstrapped Opportunistic Adversarial Curriculum Learning (BCL)}, a novel flexible adversarial curriculum learning framework for robust reinforcement learning.
Our framework combines two ideas: conservatively bootstrapping each curriculum phase with highest quality solutions obtained from multiple runs of the previous phase, and opportunistically skipping forward in the curriculum.
In our experiments we show that the proposed BCL framework enables dramatic improvements in robustness of learned policies to adversarial perturbations.
The greatest improvement is for Pong, where our framework yields robustness to perturbations of up to 25/255; in contrast, the best existing approach can only tolerate adversarial noise up to 5/255. Our code is available at: \url{https://github.com/jlwu002/BCL}.
\end{abstract}
\section{Introduction}

Advances in reinforcement learning coupled with state of the art deep neural network-based representations have led to breakthroughs in a broad range of applications, including the AlphaZero general game-playing approach~\cite{Silver18}, autonomous driving~\cite{Kiran21}, navigation of stratospheric baloons~\cite{Bellemare20}, medical imaging~\cite{Zhou21deep}, and many others.
However, a series of recent efforts demonstrated that policies learned by deep reinforcement learning (DRL) can be extremely fragile to small adversarial perturbations to input state observations~\cite{Lin17,Sun20,Wu21,Zhang2021}.
Indeed, this echoes a broader pattern of fragility of neural network architectures to adversarial perturbation attacks~\cite{Athalye2018SynthesizingRA,Eykholt2018RobustPA,Carlini2017,Goodfellow15,Szegedy14,Vorobeychik2018book}.

In turn, a series of efforts have emerged aiming to improve robustness of deep neural networks for supervised learning~\cite{Cai18,Cohen2019CertifiedAR,Raghunathan18,Goodfellow15,madry2018towards,Vorobeychik2018book}, as well as deep reinforcement learning~\cite{Oikarinen2021,Zhang2020,Zhang2021}.
However, while variations of adversarial training have proved relatively successful at attaining robustness of deep neural networks in supervised settings, success has been more modest in reinforcement learning, where the best approaches can  tolerate only very small-magnitude perturbations (e.g., up to 5/255 in Pong, which was achieved in \citet{Oikarinen2021} through RADIAL-A3C training).

We propose a novel curriculum learning framework, \emph{Bootstrapped Opportunistic Adversarial Curriculum Learning (BCL)} to boost robustness of DRL.
Our approach is inspired by recent successful curriculum learning approaches in adversarial supervised learning~\cite{Balaji19,Cai18,Sitawarin21}, but also differs substantively from these.
In particular, both \citet{Balaji19} and \citet{Cai18} propose to construct a simple curriculum of increasing input difficulty; this is what we call \emph{naive curriculum learning} below, and we show that it is not particularly effective in achieving robustness in DRL.
\citet{Sitawarin21} propose an adaptive curriculum by customizing difficulty to specific inputs as a function of attack success on each input.
This idea, however, is not meaningful in DRL, where inputs are states and success is measured in terms of overall reward of a policy, rather than accuracy of predictions on individual inputs.
Our approach is also inspired by recent success of curriculum learning approaches in reinforcement learning~\citep{narvekar2017autonomous,narvekar2020curriculum}; however, ours is the first curriculum learning framework for \emph{adversarial} reinforcement learning.

In the proposed BCL framework, we leverage two key ideas.
First, we bootstrap each phase of the curriculum by ensuring that the result of the previous phase is successful, which we do by choosing the best result over multiple adversarial training runs.
Second, we introduce adaptivity by opportunistically skipping forward in the curriculum if we find that the model learned in the current phase is already robust to the adversarial perturbations with higher magnitude.

We evaluate the efficacy of the proposed BCL framework in boosting robustness of DQN-style approaches with minimal reduction in nominal (non-adversarial) reward through extensive experiments on the Pong, Freeway, BankHeist, and RoadRunner OpenAI domains.
In all cases, we show that BCL yields considerable improvements in robustness compared to the state of the art.
In Pong, BCL-trained policies achieve near-flawless performance under adversarial perturbations of up to $\epsilon = 25/255$; in comparison, the state-of-the-art RADIAL-DQN performs poorly even with $\epsilon = 5/255$ (the reward under 30-step PGD attack is -17.7).
In BankHeist, BCL training achieves an  order-of-magnitude higher robustness for $\epsilon=15/255$ than state of the art (SA-DQN), and for RoadRunner, it is several orders of magnitude better (RADIAL-DQN is state of the art).

In summary, we make the following contributions:
\begin{enumerate}
    \item A novel flexible adversarial curriculum learning framework for reinforcement learning (BCL), in which bootstrapping each phase from multiple executions of previous phase plays a key role,
    \item A novel opportunistic adaptive generation variant that opportunistically skips forward in the curriculum,
    \item An approach that composes interval bound propagation and FGSM-based adversarial input generation as a part of adaptive curriculum generation, and
    \item An extensive experimental evaluation using OpenAI Gym Atari games (main paper) and Procgen (Appendix \ref{app:ppo_main_results}) that demonstrates significant improvement in robustness due to the proposed BCL framework.
\end{enumerate}
\section{Related Work}

Robustness to adversarial perturbations has been a subject of considerable attention in machine learning broadly, although much of the focus, and the most significant progress, has been specifically in supervised learning~\cite{Athalye2018SynthesizingRA,Eykholt2018RobustPA,Carlini2017,Goodfellow15,Szegedy14,Vorobeychik2018book}.
In particular, in the supervised learning settings, adversarial training has emerged as a major paradigm for enhancing robustness~\cite{Cai18,Cohen2019CertifiedAR,Raghunathan18,Goodfellow15,madry2018towards,Tong19,Vorobeychik2018book,Wu20}.

Studies of adversarial state perturbations to policies learned using deep  reinforcement learning are somewhat more recent~\cite{Behzadan17,Kos17,Pattanaik18,Wu21}, as are approaches for increasing robustness~\cite{Akkaya19,Fortunato17,Oikarinen2021,Pattanaik18,Tobin17,Zhang2020,Zhang2021}.
Adversarial training techniques, using either lower or upper bounds on adversarial loss have been explored, but the efficacy of conventional adversarial training has been somewhat limited, with success restricted to weak FGSM attacks, or relatively small-size perturbations~\cite{Behzadan17,Kos17,Pattanaik18}.
A number of heuristic techniques, such as adding noise at training, have also been proposed~\cite{Akkaya19,Fortunato17,Tobin17}, but these are generally not as effective against strong attacks as those based on adversarial training.
An orthogonal idea that attempts to introduce robustness directly at decision time is CARRL~\cite{Everett21}.
However, their reliance on linear bounds makes it only suitable for low-dimensional settings~\cite{Weng18}.
Among the most recent and most effective of approaches based on forms of adversarial training are RADIAL~\cite{Oikarinen2021} and SA-DQN~\cite{Zhang2020}, and we compare to these directly.
Finally, CROP is a recent approach for certifying robustness of deep reinforcement learning methods~\cite{Wu22}.
However, CROP is not in itself a method for \emph{improving} DRL robustness, either empirical or certified.

Our approach builds on prior work on the use of curriculum learning in adversarial settings~\cite{Balaji19,Cai18,Sitawarin21}, as well as curriculum learning in supervised~\citep{Bengio09} and reinforcement~\citep{narvekar2017autonomous,narvekar2020curriculum} learning.
However, as elaborated in the introduction, ours is the first \emph{adversarial} curriculum learning framework in the reinforcement learning context with a particular attention to how to design a curriculum; prior approaches for adversarial curriculum learning either do not consider a curriculum design question, or are not applicable in reinforcement learning where efficacy depends on the process dynamics and cannot be evaluated independently for each input.

\section{Preliminaries}
In this section we introduce the basics of deep reinforcement learning (DRL), focusing primarily on Deep Q-learning that we leverage in the proposed \Alg~framework.

\subsection{Deep Reinforcement Learning}

Reinforcement learning models the world as a Markov Decision Process (MDP).
An MDP is a tuple $(\mathcal{S}, \mathcal{A}, P, R, \gamma)$, where $\mathcal{S}$ is the state space, $\mathcal{A}$ is the action space, $P(s'|s,a)$ the (in our setting, unknown) transition function that determines the distribution of the next state $s'$ given current state $s$ and action 
$a$, and $R(s,a)$ the expected reward function obtained from taking action $a$ in state $s$.
Finally, $\gamma \in [0,1)$ is the temporal discount factor.
Solving MDPs amounts to computing either the Q function, $Q(s,a)$, which is the maximum discounted sum of rewards that can be achieved starting in state $s$ and taking an action $a$, or the value function $V(s) = \max_a Q(s,a)$.
A solution to an MDP is a policy $\pi(s) \in \arg\max_a Q(s,a)$.

In \emph{deep reinforcement learning (DRL)}, a key step is to approximate the value function, Q function, and/or policy using a deep neural network.
Algorithms differ both in which of these they approximate, and the particular ways these are learned from experience.
We focus on \emph{Deep Q-Network (DQN)}, a class of approaches that learn a parametric representation of the Q function.

Specifically, DQN approximates the Q function using a deep neural network $Q(s,a;\theta)$ with parameters $\theta$.
A basic DQN learning algorithm learns $Q(s,a;\theta)$ by using the loss function 
\[\mathcal{L}(\theta)=\mathbb{E}_{\left(s, a, s^{\prime}, r\right)}\left[\left(r+\gamma \max _{a^{\prime}} Q\left(s^{\prime}, a^{\prime} ; \theta\right)-Q(s, a ; \theta)\right)^{2}\right].\]

We make use of several improvements on DQN: Double DQN~\cite{VanHasselt2016} and Dueling DQN~\cite{Wang2016}. 
Double DQN uses two Q-networks with $Q_{\text{target}}$ for evaluation and $Q_{\text{actor}}$ for training, with the loss function
\begin{align}
    \mathcal{L} \left(\theta_{\text {actor}}\right)&=\mathbb{E}_{\left(s, a, s^{\prime}, r\right)}\left[\left(r+\gamma \max _{a^{\prime}} Q_{\text {target}}(s^{\prime}, a^{\prime};\theta_{\text {target}})\right.\right.\nonumber \\&\left.\left.-Q_{\text {actor}}(s, a ;\theta_{\text {actor}})\right)^{2}\right].\label{eq:standard_loss}
\end{align}
Dueling DQN is based on Double DQN and uses two estimators, one for state value function estimation (i.e., $V_Q(s)$), and one for the state-dependent action advantage function estimation $A_{Q(s,a)}$, with $Q(s,a) = V_Q(s) + A_{Q(s,a)}$.

\subsection{Adversarial Deep RL}
\label{S:advrl}

\paragraph{Adversarial Policy Perturbations} 

In adversarial perturbation attacks on DRL, 
an adversary adds a perturbation $\delta$ to each observed state $s$ constrained to be $\|\delta\|_p \le \epsilon$ (for exogenously specified $l_p$ norm and $\epsilon$) so as to minimize expected discounted reward of the executed policy $\pi(s)$.
We take $p=\infty$ here, as is common.
If the policy is based on maximizing the learned Q function, as in DQN and its variants, the attack aims to perturb this function, indirectly affecting the policy, while if DRL is based on policy learning (e.g., actor-critic), with the policy itself represented by a neural network $\pi(s;\theta)$, the policy is attacked directly.
Specifically, a common attack on DQN aims maximize $\mathcal{L}(\mathrm{Softmax}(Q(s+\delta;\theta)),\pi(s))$ with respect to $\delta$, where $\mathcal{L}$ is the cross-entropy loss, $Q(s)$ is the vector of Q values over all actions in state $s$.
A PGD (projected gradient descent) attack~\cite{madry2018towards} is then implemented with this loss function, which updates $\delta$ iteratively:
$\delta_{k+1} \leftarrow \delta_k + \alpha \cdot \textrm{sign} (\nabla_\delta \mathcal{L}(Q(x + \delta_k;\theta), \pi(s)))$ over a fixed number of iterations, projecting to a nearest feasible state and clipping to ensure that $\|\delta\|_\infty \le \epsilon$.
In policy learning methods, a common loss function is instead $\mathcal{L}(\pi(s+\delta;\theta), \pi(s))$, with PGD attacks implemented just as above.
An important special class of PGD is FGSM (fast gradient sign method)~\cite{Goodfellow15}, in which PGD is executed for only a single  iteraration and $\alpha=\epsilon$.

\paragraph{Adversarial Training}

Deep RL is robust to adversarial policy perturbations with magnitude up to $\epsilon$ if attacks do not significantly reduce the discounted sum of rewards.
Common approaches aimed at robust learning in general use some form of adversarial training, where after initially training the model in the regular manner, additional training phases either add adversarial perturbations to inputs that are used in further gradient updates~\cite{madry2018towards}, or take gradients of an upper bound on adversarial loss~\cite{Wong18b}.
A state-of-the-art form of adversarial training uses PGD attacks to generate adversarial perturbations~\cite{madry2018towards}.
A recent alternative which is much more computationally efficient and  equally efficacious uses FGSM with random initializations instead~\cite{Wong2020};
henceforth, we term this variant RI-FGSM.

In robust DQN, a recent RADIAL-DQN approach~\citep{Oikarinen2021} on which we build defines the loss function as
\begin{equation}
    \mathcal{L}_{\text{RADIAL}}=\kappa \mathcal{L}_{\text{standard}}+(1-\kappa) \mathcal{L}_{\text{adv}},
    \label{eq:radial_loss}
\end{equation}
where  $\mathcal{L}_{\text {standard}}$ is defined in Equation \eqref{eq:standard_loss}, and
\begin{align*}
\begin{split}
    \mathcal{L}_{\text{adv}}\left(\theta_{\text {actor}}, \epsilon\right)=\mathbb{E}_{\left(s, a, s', r\right)} \left[
    \sum_y \mathcal{L}_y(s,a)\right],
    \end{split}
\end{align*}
with 
\(
\mathcal{L}_y(s,a)=\left(r+\gamma \max_{a^{\prime}} Q_{\text{target}}\left(s^{\prime}, a^{\prime}\right)-\tilde{Q}^{\epsilon}_{\text{actor}}(s, y)\right)^{2}
\)
when $y=a$ and
\(
\mathcal{L}_y(s,a) = \left(Q_{\text{actor}}(s, y)-\tilde{Q}^{\epsilon}_{\text{actor}}(s, y)\right)^{2}
\)
otherwise.
This is referred to as approach \# 1 in \citet{Oikarinen2021}, which yields
 a strict upper bound on the loss function under adversarial perturbation, that is, $\mathcal{L}_{\text{standard}}(s+\delta;\epsilon) \leq \mathcal{L}_{\text{adv}}(s; \epsilon)$ with $||\delta||_p \leq \epsilon$; approach \# 2 is an alternative that aims to minimize the weighted overlapped IBP Q-values. Since approach \# 2 empirically outperforms approach \# 1 for RADIAL-DQN, we use approach \# 2 in our experiments below for RADIAL curriculum training.
RADIAL-DQN (both approaches) uses $\tilde{Q}^{\epsilon}_{\text{actor}}(s, y)$ which is an upper or lower bound on $Q_{\text{actor}}(s, y)$ derived using interval bound propagation (IBP) for a given attack budget $\epsilon$.
As IBP yields relatively loose bounds, $\epsilon$ must of necessity be small for these to be meaningful, limiting the ability to achieve robustness beyond relatively low values of $\epsilon$.

\section{Bootstrapped Opportunistic Adversarial Curriculum Learning}
\label{S:bcl}

Curriculum learning is an old concept in iterative learning in which easier examples are provided before more challenging examples~\cite{Bengio09}.
A natural alternative is to start with small values of $\epsilon$ and gradually increase these during adversarial training.
However, as we show in the experiments, this latter idea works extremely poorly for DRL.
We propose a novel \emph{Bootstrapped Opportunistic Adversarial Curriculum Learning (BCL)} framework for iterative adversarial training.
The key idea is to bootstrap each training step to ensure that subsequent iterations begin with a partially robust baseline, and to also enable the algorithm to ``skip forward'' if robustness against several successive values of $\epsilon$ has already been achieved.
Our BCL framework allows one to explicitly trade off between being conservative (paying more attention to the former) and opportunistic (greater focus on the latter).

\subsection{The BCL Algorithm}
\label{BCL_Algo}
At the high level, the proposed BCL algorithmic framework begins by creating a \emph{baseline curriculum}, that is, an increasing sequence of $L$ attack budgets $\{\epsilon_i\}$, with $\epsilon_1 < \epsilon_2 < \cdots < \epsilon_L$, where $\epsilon_L=\epsilon$ is our target robustness level.
It also begins with a sufficiently small $\epsilon_0 > 0$ so that it is either already achievable (e.g., by standard DRL, prior art such as RADIAL, etc) or not difficult to achieve; we assume that BCL is initially \emph{bootstrapped} with a model $f_{\theta_0}$ that is indeed able to achieve this relatively low bar.
It then proceeds through a series of phases, where a phase is associated with attempting to achieve robustness against $\epsilon_i$ in the curriculum for some $i$ (which is not necessarily identical to the phase number, as we discuss below).
In each phase, we run adversarial training (AT) up to $K$ times, where each AT run is bootstrapped by the best model obtained thus far, $f_\theta$.
Each model thereby learned is then independently evaluated, and if the best model obtained thus far in the current phase exhibits sufficiently good performance (a criterion for this can depend on $\epsilon_i$, and represented by a function $\overline{V}(\epsilon)$ in Algorithm~\ref{alg:cl}), we can stop and move to the next phase as long as we performed at least a minimum number $K_{\mathrm{min}}$ AT runs.
The best model in the current phase then becomes the best model achieved thus far, updating $f_\theta$.
Algorithm~\ref{alg:cl} describes this procedure more precisely.

\begin{algorithm}[h]
   \caption{BCL algorithm.}
   \label{alg:cl}
\begin{algorithmic}
   \STATE {\bfseries Input:} 
   $\epsilon$, $K$, $K_{\mathrm{min}}$, $\overline{V}(\epsilon)$, $f_{\theta_0}$.
   \STATE $f_\theta \leftarrow f_{\theta_0}$ \quad // Initialization
   \STATE $\{\epsilon_i\}_{i=1}^L \leftarrow \mathrm{Curriculum}(\epsilon)$ // Create curriculum
   \STATE $(i,\epsilon_{\mathrm{best}}) \leftarrow \mathrm{ChooseNext}(f_\theta,\{\epsilon_i\},0,\overline{V}(\epsilon))$
   \WHILE{$\epsilon_{\mathrm{best}} < \epsilon$}
   \FOR{$k = 1,\ldots, K$}
   \STATE $f_{\theta_k} \leftarrow \mathrm{Train}(f_\theta,\epsilon_i)$
   \STATE $V_k \leftarrow \mathrm{Eval}(f_{\theta_k},\epsilon_i)$
   \IF{$k \ge K_{\mathrm{min}}$ and $V_k \ge \overline{V}(\epsilon_i)$}
   \STATE {\bfseries break}
   \ENDIF
   \ENDFOR
   \STATE // Find  the best model among training results
   \STATE $k^{*}\leftarrow \arg\max_{k\in [K]} V_k$ %
   \STATE $f_\theta \leftarrow f_{\theta_{k^*}}$
   \STATE $(i,\epsilon_{\mathrm{best}}) \leftarrow \mathrm{ChooseNext}(f_\theta,\{\epsilon_i\},i,\overline{V}(\epsilon))$
   \ENDWHILE
   \STATE {\bfseries return} {$f_\theta$} 
\end{algorithmic}
\end{algorithm}

The next central feature of BCL is the ability to \emph{skip forward} in the curriculum, omitting the next budget level $\epsilon_{i+1}$, and potentially others after it, as shown in the $\mathrm{ChooseNext}$ step (Algorithm~\ref{alg:next}).
The most we can skip forward is to the smallest $\epsilon_j$ to which the current model is \emph{not robust} (this is the purpose of $\mathrm{EvalRobust}$ function in Algorithm~\ref{alg:next}).
This skipping feature is most useful because it significantly reduces the time that BCL needs to run, but as we show in the experiments, there are times where it also yields better robustness than obtained by following the baseline curriculum.

\begin{algorithm}[h]
   \caption{ChooseNext}
   \label{alg:next}
\begin{algorithmic}
   \STATE {\bfseries Input:} 
   $f_\theta,\{\epsilon_i\}_{i=1}^L,j,\overline{V}(\epsilon)$.
    \STATE // Find smallest $i$ such that $f_\theta$ is \emph{not} robust for $\epsilon_i$
   \STATE $i \leftarrow \mathrm{EvalRobust}(f_\theta,\{j+1,\ldots,L\},\overline{V}(\epsilon))$
   \STATE // Select index $l$ to train with next
   \STATE $l \leftarrow \mathrm{Select}(\{j+1,\ldots,i\})$
   \STATE {\bfseries return} {$(l,\epsilon_{l-1})$} 
\end{algorithmic}
\end{algorithm}

Algorithm~\ref{alg:cl} takes as input a fixed target $\epsilon$ that we wish to induce robustness to, but in practice it is often the case that we wish to be more opportunistic, and simply observe what is possible in trading off robustness and baseline (non-adversarial) efficacy.
For example, we can set $\epsilon$ to be very high, but stop BCL well in advance of reaching it if we observe significant performance degradation.

For the RADIAL curriculum training, unlike  training with adversarial examples (\cref{gen_adv_pert}), it does not have a target $\epsilon$ to be robust against. We choose to always follow the baseline curriculum for the $\epsilon$. We find that RADIAL training does not increase the nominal reward in trend, and many times the significant decrease in nominal reward is accompanied by the decease in robustness as the model begin to collapse.  Thus, we aim at maintaining the nominal reward at a high level. We set the a threshold for each model, and re-train the model for maximum $K$ times if the nominal reward is below the threshold. We stop the training if nominal reward is below the threshold level for $M$ consecutive curriculum phases.

Next, we illustrate the BCL framework with several special cases, noting first that both conventional adversarial training and naive curriculum learning can also be viewed as variants of BCL.

\noindent{\bf Adversarial Training (AT): } Standard adversarial training can be viewed as a special case of BCL if $K=1$ and the baseline curriculum is simply the singleton $\epsilon$.

\noindent{\bf Naive Curriculum Learning (NCL): } Setting $K=1$ and always following the baseline curriculum (i.e., the next index returned by the $\mathrm{ChooseNext}$ function is always ${i+1}$) recovers a naive implementation of curriculum learning.

\noindent{\bf Conservatively Bootstrapped Curriculum Learning (BCL-C): } If we set $K_{\mathrm{min}} = K$ and always follow the baseline curriculum, BCL never opportunistically skips forward, and setting $K$ sufficiently high ensures that  each step is bootstrapped with an effective model trained using all smaller values of $\epsilon$.

\noindent{\bf Maximum Opportunistic Skipping (BCL-MOS): } If we always choose to skip to the smallest $\epsilon$ against which the current model $f_\theta$ is not (yet) robust, we obtain the most opportunistic version of the algorithm.

\subsection{Generating Adversarial Perturbations}
\label{gen_adv_pert}
The key question left open in BCL is precisely how we train a model in a particular phase to be robust against a given adversarial budget $\epsilon$.
There are two major ways to do this: using bounds on the impact of adversarial perturbations, such as those produced by IBP, as done by RADIAL~\cite{Oikarinen2021}, and using adversarial perturbations~\cite{Zhang2020,Zhang2021}.
In addition to using IBP, RADIAL introduces a crucial insight in robustness training in distinguishing updates for actions that have been chosen (for which the immediate reward has been observed) from those that have not been, as discussed in Section~\ref{S:advrl}.
We leverage this idea, but replace IBP with adversarial examples.
Next, we present a novel approach for generating adversarial examples for adversarial training in each phase of BCL that specifically leverages DDQN.

Recall that in RADIAL-DQN (approach \# 1), $\tilde{Q}^\epsilon_{\mathrm{actor}}(s,a)$ uses IBP bounds on the Q function that can be achieved through adversarial perturbations.
Alternatively, we can define it as $\tilde{Q}^\epsilon_{\mathrm{actor}}(s,a) = Q_{\mathrm{actor}}(s+\delta^*,a)$, where $\delta^*$ (approximately) solves the following optimization problem:
\begin{align}
\label{eq:DQN_opt_pert}
\min_{||\delta||_\infty \leq \epsilon}\sum_{a \in A} \pi(s+\delta,a) Q_{\text{target}}(s,a),
\end{align}
 where $\epsilon$ is the bound on $l_\infty$-norm of the perturbation (as is common in prior literature on robust reinforcement learning), with $\pi(s+\delta,a) = 1$ iff $a$ is the best action to be taken after observing $s+\delta$, i.e.,
 \(
 a \in \arg\max_{a'} Q_{\text{actor}}(s+\delta,a'), 
 \)
 and $\pi(s+\delta,a) = 0$ otherwise.
  In other words, Equation \eqref{eq:DQN_opt_pert} aims to identify $\delta$ that minimizes the expected discounted sum of rewards as approximated by $Q_{\text{target}}(s,a)$.
  Note that here it is crucial to separate the $Q_{\text{actor}}$, which determines the policy, and $Q_{\text{target}}$, which serves as an ``objective'' evaluation of state-action values.
  This is in contrast with typical adversarial perturbation attacks on DRL described in Section~\ref{S:advrl}, where the adversary merely aims to prevent a target (optimal) action from being chosen, but may well incentivize DRL to choose a near-optimal action instead.
  
In order to approximately solve the problem in Equation~\eqref{eq:DQN_opt_pert}, we first replace a policy $\pi$ by its differentiable approximation $\tilde{\pi}$, where 
\[
\tilde{\pi}(s,a) = \frac{e^{Q_{\text{actor}}(s,a)}}{\sum_{a'} e^{Q_{\text{actor}}(s,a')}}.
\]
Equivalently, $\tilde{\pi}(s+\delta) = \mathrm{Softmax} (Q_{\text{actor}}(s + \delta))$,
where we use $Q(s)$ to denote a vector with values for each action $a$.
We then solve the following proxy optimization problem to approximate $\delta^*$:
\begin{equation}
    \min_{||\delta||_\infty \leq \epsilon}
    \mathrm{Softmax} (Q_{\text{actor}}(s + \delta))\odot Q_{\text{target}}(s),
    \label{eq:DQN_pert}
\end{equation}
with $\odot$ denoting the dot-product.

Commonly, the problem in Equation~\eqref{eq:DQN_pert} is solved using PGD~\cite{madry2018towards}.
However, this becomes a major bottleneck in training, particularly when we use a large number of PGD iterations.
We make two improvements to significantly reduce the  time associated with computing $\delta$.
First, we use FGSM + Random Initialization (RI-FGSM) (see Section~\ref{S:advrl}), proposed by \citet{Wong2020} for supervised adversarial training, for which it was shown highly effective.
Ours is the first application of this idea in robust DRL.
Second, we dynamically calculate the perturbation $\delta$ 
and push the entire tuple $(s, a, s', r, \delta)$ (i.e., including $\delta$) into the replay buffer. This enables us to re-utilize the previously calculated perturbations to further improve training efficiency. Consequently, we chose a relatively small replay buffer size to ensure the perturbations $\delta$ stored in the buffer are frequently updated as the DRL model evolves.

With RADIAL-DQN (approach \# 2), which minimize the weighted overlapping IBP Q-values, as well as the approach above for generating \emph{specific} adversarial perturbations, which yields a lower bound on adversarial loss, we have two specific ways that we can use to compute gradient updates in the $\mathrm{Train}$ step of BCL for a given perturbation magnitude $\epsilon$.
We refer to the former simply as RADIAL, and to the latter as AT (for adversarial training).
Both can be ``plugged in'' to any variant of BCL.
Additionally, we can \emph{compose} these approaches, giving rise to a novel variant:

\noindent{\bf RADIAL + AT Bootstrapped Curriculum Learning (BCL-RADIAL+AT): } First, run BCL-RADIAL until it reaches a point in the curriculum at which its performance degrades significantly; then, switch to BCL-X-AT (where X is either C or MOS) for the remainder of the curriculum.

\section{Experiments}

\subsection{Experiment Setup}
We evaluate the proposed approach using four Atari-2600 games  from the OpenAI Gym~\cite{bellemare2013arcade}: Pong, Freeway, BankHeist, and RoadRunner. Those environments have discrete action space.
The walltime for all experiments are documented in the Appendix \ref{app:dqn_walltime}.
We use $R_{\text{nominal}}$ to denote a model's nominal reward (i.e., average discounted sum of per-step rewards without adversarial perturbations), and $R_{\text{adv}}^{\epsilon}$ to represent a model's reward under adversarial attacks with $l_\infty$ perturbation bounded by $\epsilon$. 
For each model we calculate a score 
using $R_{\text{nominal}} + \frac{1}{3}\sum_\epsilon R_{\text{adv}}^{\epsilon}$ for all $\epsilon$ listed in Table \ref{bcl_main_table} to measure the model's robustness level, and this score is used to choose the median and best final result (out of three independent runs); we present the median here, and the results of all runs, as well as the best are provided in the Appendix \ref{app:dqn_additional}.
We experiment all the BCL variations in \cref{BCL_Algo}. 
We compare \Alg-based approaches to six benchmarks: 1) standard Dueling DQN training (DQN (Vanilla)), 2) SA-DQN using convex relaxation (SA-DQN (Convex))~\cite{Zhang2020}, 3) RADIAL-DQN~\cite{Oikarinen2021}, 4) standard adversarial training (AT-DQN)~\cite{madry2018towards}, 5) naive curriculum learning with adversarial examples (NCL-AT-DQN)~\cite{Cai18,Sitawarin21} and 6) naive curriculum learning with RADIAL method (NCL-RADIAL-DQN). 
For DQN (Vanilla) we use the results from~\citet{Zhang2020}, and for AT-DQN, NCL-AT-DQN as well as NCL-RADIAL-DQN we perform our own training as three restricted variants of the BCL algorithm. The AT method is the one we purposed in \cref{gen_adv_pert}. The adversarial examples for all games are generated using RI-FGSM.

\noindent\textbf{DQN Hyperparameters }
Our implementation is based on RADIAL-DQN~\cite{Oikarinen2021}. For most hyperparameters we keep them the same as in RADIAL-DQN, with a few exceptions such as replay initial and replay buffer size, which are modified according to our model setting to improve training efficiency. We use buffer size 50,000 across all environments compared to 200,000 used by RADIAL. For replay initial we use 256 compared to 50,000 in RADIAL. We use RI-FGSM (Algorithm 3 in~\citet{Wong2020}, see Section~\ref{S:advrl}) with hyperparameter $\alpha = 0.375$ for approximating $\delta$ during training. The detailed DQN specific hyperparameters for AT runs are in Table \ref{app:tab_dqn_params}. The one exception is BCL-RADIAL+AT-DQN for RoadRunner environment: for the AT training we use $1.25 \times 10^{-7}$ as the learning rate, as we find with learning rate 0.000125 the nominal reward would decrease significantly after training. For NCL/BCL-RADIAL-DQN, all hyperparameters are the same as in RADIAL-DQN.

To ensure a fair comparison, we let all methods to have the same computational constraints and evaluation metrics: for all environments we train for 4.5 million frames (same as RADIAL-DQN) for each run,  evaluate over 20 test episodes and report the averaged reward.

\begin{table}[h]
\caption{DQN specific hyperparameters (AT runs)}
\label{app:tab_dqn_params}
\vskip 0.15in
\begin{center}
\begin{small}
\begin{sc}
\begin{tabular}{ll}
\toprule
Parameter & Value \\\midrule
Discount Factor ($\gamma$)                   & 0.99           \\
Buffer Size             & 50000          \\
Replay Initial          & 256            \\
Batch Size              & 128            \\
Optimizer               & Adam \\
Optimizer Learning Rate & 0.000125    \\
\bottomrule
\end{tabular}
\end{sc}
\end{small}
\end{center}
\vskip -0.1in
\end{table} 

\noindent\textbf{Adversarial Attacks for DQN}
As we observe significant issues with obfuscated gradients with NCL/BCL-RADIAL-DQN, we apply four types of adversarial attacks for DQN models: 1) 30-step untargeted PGD attack with step size 0.1 (this is stronger than the 10-step PGD used in \citet{Oikarinen2021}); 2) RI-FGSM ($\alpha = 0.375$); 3) RI-FGSM (Multi): sample $N = 1000$ random starts for RI-FGSM, and takes the first sample where the resulting adversarial example alters the action; 4) RI-FGSM (Multi-T): sample $N = 1000$ random starts for RI-FGSM, and takes the sample which results the agent taking the action corresponding to the lowest Q value among those $N$ samples. 
We report the lowest reward obtained after running those four attacks. We observe that with obfuscated gradients, RI-FGSM (Multi-T) results in the strongest attack in many cases, while 30-step PGD is typically stronger otherwise (see the Appendix \ref{app:dqn_additional} for details).

\noindent\textbf{Hyperparameters for AT-DQN and NCL-AT-DQN} For AT-DQN, we experiment with a series of varying values of $\epsilon$, and present the most effective results, with the comprehensive results deferred to Appendix \ref{app:dqn_additional}. We generate adversarial perturbations as in \cref{gen_adv_pert}, and use DQN (Vanilla) as $f_{\theta_0}$,\footnote{For the RoadRunner environment we used the implementation of vanilla DQN from RADIAL-DQN (version 1) as $f_{\theta_0}$, which yields better results.} setting $K = L = 1$. For each environment we only need one run which is 4.5 million frames. 

For naive curriculum learning ({NCL-AT-DQN}), we use DQN (Vanilla) as $f_{\theta_0}$, as for AT-DQN. We set  $K= 1$ and $\epsilon_0 = 0$, with curriculum increment of 1/255 (i.e., using the baseline curriculum) until target $\epsilon$ is reached. The choice of $\epsilon$ of each environment is the same as the ones in BCL experiments. 
However, because NCL-AT-DQN significantly underperforms BCL and (unlike BCL) when the target $\epsilon$ is reached the final NCL-AT-DQN model performs extremely poorly, we instead report the best result along the curriculum path to provide the strongest benchmark. 

\noindent\textbf{Hyperparameters for NCL/BCL-RADIAL-DQN} In NCL/BCL-RADIAL-DQN experiments, we bootstrap from RADIAL-DQN. We set $\epsilon_0 = 1/255$, with $\epsilon$ always follows the baseline curriculum. The increments of the baseline curriculum is $1/255$. We set $K = 1$ for NCL-RADIAL-DQN and $K = 3$ for BCL-RADIAL-DQN. We perform maximum $K$ runs for each curriculum phase; if none of the $K$ run results has  a nominal reward  above the threshold, we choose the one with the highest nominal reward and move to the next curriculum phase. We stop the training if nominal reward is below the threshold for $M = 2$ consecutive curriculum phases.  The thresholds are shown in Table \ref{TH_RADIAL_DQN} in Appendix \ref{app:thresholds_dqn}. We report the best result along the curriculum path for  NCL/BCL-RADIAL-DQN.

\noindent\textbf{Hyperparamters for BCL-RADIAL+AT-DQN} For BCL-RADIAL+AT-DQN, we choose the median run among three BCL-RADIAL-DQN runs and perform BCL-C-AT-DQN starting with this run. 
As we are only able to perform BCL-C-AT-DQN further for BankHeist and RoadRunner, we choose $K=3$ for BankHeist, and $K = 1$ for RoadRunner. The baseline curriculum for  BankHeist starts from $\epsilon_0 = 13/255$ and for RoadRunner $\epsilon_0 = 12/255$, with curriculum increment of $1/255$, and target $\epsilon = 15/255$. 

\noindent\textbf{Other \Alg~Hyperparameters} In \Alg, we also evaluate two concrete novel instantiations of the proposed \Alg~framework: conservatively bootstrapped curriculum learning (\Alg-C-AT-DQN), and maximum opportunistic skipping (\Alg-MOS-AT-DQN). 
We bootstrapped all instances by using  RADIAL-DQN as $f_{\theta_0}$.\footnote{For the RoadRunner environment we used the version 1 implementation of RADIAL-DQN~\cite{Oikarinen2021} as $f_{\theta_0}$, which yields better bootstrapping performance. 
Nevertheless, we always use the best-performing version of RADIAL-DQN (version 2) as the benchmark in Table \ref{bcl_main_table}.
}
Further, the baseline curriculum is created starting with $\epsilon_0 = 3/255$, since RADIAL-DQN (which we use as $f_{\theta_0}$) is already robust up to $3/255$. 
The baseline curriculum is then created by using increments of $1/255$
until reaching the target $\epsilon$.
The BCL hyperparameters ($K$, $K_{\text{min}}$ and $\epsilon$) are listed in Table \ref{tab_bcl_params} in the Appendix \ref{app:thresholds_dqn}. The thresholds $\overline{V}(\epsilon)$ for \Alg-MOS-AT-DQN are listed in Table \ref{mos_th} in the Appendix \ref{app:thresholds_dqn}, where $\overline{V}_{\text{nominal}}(\epsilon)$ is the threshold for nominal reward, and $\overline{V}_{\text{adv}}(\epsilon)$ is the threshold for rewards under adversarial attacks. As described in Section~\ref{S:bcl}, if the model in phase $i$ is trained against $\epsilon_i$ and $R_{\text{nominal}}\geq \overline{V}_{\text{nominal}}(\epsilon)$, we perform evaluation with adversarial attacks, find the maximum $j>i$ such that $R_{\text{adv}}^{\epsilon_j}\geq \overline{V}_{\text{adv}}(\epsilon)$, and skip forward in the baseline curriculum, training with $\epsilon_{j+1}$ in the next phase.
In our implementation of \Alg, we further smoothed the curriculum  by gradually increasing the upper bound $\epsilon$ on adversarial perturbations from $\epsilon_i$ to $\epsilon_{i+1}$ in phase $i+1$ during the 4.5 million training frames. The function evaluating the quality of intermediate results in Algorithm~\ref{alg:cl},
$\mathrm{Eval}(f_{\theta_k},\epsilon_i)$, returns the efficacy score $V_k = R_{\text{nominal}} + \frac{1}{2}(R_{\text{adv}}^{\epsilon_i}+R_{\text{adv}}^{\epsilon_{i-1}})$, which allows us to choose the best model among all the intermediate results. 
Note that it is crucial to include $R_{\text{nominal}}$ as a part of the criterion for model selection, as a model with a  high nominal reward tends to show considerably better stability in subsequent curriculum training.

We used a time-varying $\kappa$ in Equation \ref{eq:radial_loss} for BCL.
Specifically, we let $\kappa$ decrease from 1 to 0.5 through the 4.5 million training frames for all experiments except
 when we use RADIAL-DQN as $f_{\theta_0}$ in Pong, Freeway and BankHeist; or when we use RADIAL-DQN (version 2) as $f_{\theta_0}$ in RoadRunner (i.e., NCL/BCL-RADIAL-DQN and BCL-RADIAL+AT-DQN).
 In these cases, $\kappa$ is set to 0.8 throughout training. 
 The choice of $\kappa=0.8$ ensured consistency with the $\kappa$ used in RADIAL-DQN \cite{Oikarinen2021}, which makes the bootstrapping process more stable.

\subsection{Results}
Our main results are presented in Table~\ref{bcl_main_table}, with extensive additional results and analysis provided in the Appendix.
\begin{table*}[htbp]
\caption{Average episode rewards $\pm$ standard error of the mean (SEM) over 20 episodes. The \colorbox{Gray}{\textcolor{red}{gray rows}} are the most robust models (selected based on score $R_{\text{nominal}} + \frac{1}{3}\sum_\epsilon R_{\text{adv}}^{\epsilon}$). \textbf{Boldface} marks the best results for each value of $\epsilon$, including $\epsilon = 0$ (nominal); we marked multiple row entries as boldface for a given $\epsilon$ if they are statistically indistinguishable (i.e., have overlapping confidence intervals).
}
\label{bcl_main_table}
\vskip 0.15in
\begin{center}
\begin{small}
\begin{sc}
\begin{tabular}{lllll}
\toprule\toprule
\multicolumn{5}{c}{\textbf{Pong}}\\
    Method/Metric            & Nominal & \multicolumn{3}{c}{30-step PGD/RI-FGSM Attack}  \\
    $\epsilon$            & 0       & 10/255 & 20/255 & 25/255 \\\toprule
DQN (Vanilla)       &    $\mathbf{21.0\pm 0.0}$  & $-21.0\pm 0.0$& $-21.0\pm 0.0$& $-21.0\pm 0.0$ \\
SA-DQN (Convex)        &    $\mathbf{21.0\pm 0.0}$ & $-21.0\pm 0.0$& $-21.0\pm 0.0$&  $-21.0\pm 0.0$   \\
RADIAL-DQN     &$\mathbf{21.0\pm 0.0}$ &$-21.0\pm 0.0$&$-21.0\pm 0.0$&  $-21.0\pm 0.0$  \\\cmidrule{1-1}
AT-DQN &$ 20.8 \pm 0.1 $ &$ 9.9 \pm 3.7 $ & $ -20.8 \pm 0.1 $ & $ -21.0 \pm 0.0 $\\
NCL-AT-DQN&$\mathbf{21.0 \pm 0.0} $&$ 20.4 \pm 0.2 $ &$ -21.0 \pm 0.0 $ & $ -21.0 \pm 0.0 $\\
NCL-RADIAL-DQN&$\mathbf{21.0 \pm 0.0} $&$-20.6 \pm 0.1$ & $-21.0 \pm 0.0$ &$ -21.0 \pm 0.0 $\\
\midrule
\rowcolor{Gray}\textcolor{red}{\Alg-C-AT-DQN} &$\mathbf{ 21.0 \pm 0.0} $ & $ \mathbf{ 21.0 \pm 0.0} $ & $ \mathbf{ 21.0 \pm 0.0} $ & $ \mathbf{ 21.0 \pm 0.0} $\\
{\Alg-MOS-AT-DQN}&$ \mathbf{ 21.0 \pm 0.0} $&$ \mathbf{ 21.0 \pm 0.0} $ & $ 20.9 \pm 0.0 $ & $ 20.9 \pm 0.0 $\\
{\Alg-RADIAL-DQN}&$\mathbf{21.0\pm 0.0}$  &$\mathbf{ 21.0 \pm 0.0} $ & $ -20.9 \pm 0.1 $ & $ -21.0 \pm 0.0 $ \\
\bottomrule \toprule
\multicolumn{5}{c}{\textbf{Freeway}}\\
    Method/Metric            & Nominal & \multicolumn{3}{c}{30-step PGD/RI-FGSM Attack}  \\
    $\epsilon$            & 0       & 10/255 & 15/255 & 20/255 \\\toprule
DQN (Vanilla)     &$\mathbf{33.9 \pm 0.1} $&$ 0.0  \pm 0.0 $&$ 0.0  \pm 0.0 $&$ 0.0  \pm 0.0 $    \\
SA-DQN (Convex)       &$30.0 \pm 0.0 $&$ 19.3 \pm 0.4 $& $19.3 \pm 0.3 $&$ 20.0 \pm 0.3 $   \\
RADIAL-DQN   &$33.2 \pm 0.2 $&$ 17.1 \pm 0.3 $ & $ 13.4 \pm 0.2 $ & $ 7.9  \pm 0.3 $   \\\cmidrule{1-1}
AT-DQN &$32.4 \pm 0.2 $&$ 0.0 \pm 0.0 $&$ 0.0 \pm 0.0 $&$ 0.0 \pm 0.0$\\
NCL-AT-DQN&$32.8 \pm 0.2 $&$22.0 \pm 0.5$ & $9.6 \pm 0.4$ & $0.0 \pm 0.0$\\
NCL-RADIAL-DQN&$33.5 \pm 0.2$&$ 9.7 \pm 0.5 $ & $ 11.6 \pm 0.5 $&$18.0 \pm 0.4$\\
\midrule
{\Alg-C-AT-DQN} &$\mathbf{34.0 \pm 0.0 }$&$28.8 \pm 0.4$ & $ 21.6 \pm 0.5 $ & $ 17.4 \pm 0.2$  \\
\rowcolor{Gray}\textcolor{red}{\Alg-MOS-AT-DQN}&$\mathbf{34.0  \pm 0.0}  $&$ 31.1  \pm 0.3  $&$ \mathbf{25.9 \pm 0.4 }$&$ \mathbf{20.8 \pm 0.3} $  \\
\rowcolor{Gray}\textcolor{red}{\Alg-RADIAL-DQN}&$33.1 \pm 0.1$&$\mathbf{33.4 \pm 0.1}$ & $\mathbf{25.9 \pm 0.6}$ & $\mathbf{ 21.2 \pm 0.5 }$ \\
\bottomrule \toprule
\multicolumn{5}{c}{\textbf{BankHeist}}\\
    Method/Metric            & Nominal & \multicolumn{3}{c}{30-step PGD/RI-FGSM Attack}  \\
    $\epsilon$            & 0       & 5/255 & 10/255 & 15/255 \\\toprule
DQN (Vanilla)     &$1325.5  \pm 5.7$      & $0.0    \pm 0.0$      & $0.0    \pm 0.0$     & $0.0    \pm 0.0$      \\
SA-DQN (Convex)       &$1237.5 \pm 1.7$   & $ 1126.0 \pm 32.0 $ & $ 63.0 \pm 3.5 $ &  $ 16.0 \pm 1.6 $\\
RADIAL-DQN   &$\mathbf{1349.5  \pm 1.7}$    & $ 581.5 \pm 16.7 $    & $0.0    \pm 0.0$     &    $0.0    \pm 0.0$   \\\cmidrule{1-1}
AT-DQN &$1271.0 \pm 15.5 $&$ 129.0 \pm 10.2 $&$ 5.5  \pm 1.1$ & $0.0  \pm 0.0$\\
NCL-AT-DQN&$1311.0 \pm 4.0  $&$ 245.0 \pm 23.7 $&$ 1.0 \pm 0.7$ &$ 0.0  \pm 0.0 $\\
NCL-RADIAL-DQN&$1272.0 \pm 10.7$  & $1168.0 \pm 3.4$ & $59.5 \pm 7.6$ & $9.0 \pm 1.9$\\
\midrule
{\Alg-C-AT-DQN} &$ 1285.5 \pm 5.2 $ &$ 1143.5 \pm 30.0 $ & $ 988.5 \pm 12.3 $ & $ 250.5 \pm 14.6 $\\
{\Alg-MOS-AT-DQN} &$ 1307.5 \pm 9.5 $&$ 1095.5 \pm 6.2 $ & $ 664.0 \pm 60.6 $ & $ 586.5 \pm 105.6 $\\
{\Alg-RADIAL-DQN}&$1225.5 \pm 4.9$ & $ \mathbf{1225.5 \pm 4.9} $ & $\mathbf{ 1223.5 \pm 4.1 }$ & $ 228.5 \pm 13.9 $ \\
\rowcolor{Gray}\textcolor{red}{\Alg-RADIAL+AT-DQN}&$1215.0 \pm 8.4$ & $1093.0 \pm 5.3$ & $1010.5 \pm 8.0$ & $\mathbf{961.5 \pm 9.2}$ \\
\bottomrule \toprule
\multicolumn{5}{c}{\textbf{RoadRunner}}\\
    Method/Metric            & Nominal & \multicolumn{3}{c}{30-step PGD/RI-FGSM Attack}  \\
    $\epsilon$            & 0       & 5/255 & 10/255 & 15/255 \\\toprule
DQN (Vanilla)     &$43390   \pm 973$        & $0     \pm 0$     & $0     \pm 0$      & $0     \pm 0$      \\
SA-DQN (Convex)       &$\mathbf{45870   \pm 1380}$    & $985   \pm 207$   & $0     \pm 0$      & $0     \pm 0$      \\
RADIAL-DQN   &$44595   \pm 1165$    & $7195  \pm 929$  & $495    \pm 116$    & $ 0 \pm 0 $     \\\cmidrule{1-1}
AT-DQN &$ 39890 \pm 2092 $ & $ 20160 \pm 1973 $ & $ 0 \pm 0 $ & $ 0 \pm 0 $\\
NCL-AT-DQN &$\mathbf{47925 \pm 1123} $&$ 37745 \pm 2014 $ & $ 10 \pm 10 $ & $ 0 \pm 0 $\\
NCL-RADIAL-DQN&$ 41045 \pm 1289 $ & ${ 37865 \pm 1082} $ & $ \mathbf{37865 \pm 1082 }$ & $ 6350 \pm 590 $ \\
\midrule
{\Alg-C-AT-DQN}&$\mathbf{45815 \pm 1422}$ &$ 31305 \pm 3590 $ & $ 11405 \pm 1385 $ & $ 6335 \pm 716 $\\
{\Alg-MOS-AT-DQN}& $ 44275 \pm 1997 $&$\mathbf{ 40060 \pm 1828} $ & $ 15785 \pm 1124 $ & $ 1195 \pm 180 $\\
{\Alg-RADIAL-DQN}&$ 41045 \pm 1289 $ & $ {37865 \pm 1082} $ & $ \mathbf{37865 \pm 1082 }$ & $ 6350 \pm 590 $ \\
\rowcolor{Gray}\textcolor{red}{\Alg-RADIAL+AT-DQN}&$ 42490 \pm 1309 $ & $\mathbf{ 42490 \pm 1309} $ & $ \mathbf{37665 \pm 1563 }$ & $\mathbf{ 25325 \pm 1057} $ \\
\bottomrule 
\end{tabular}
\end{sc}
\end{small}
\end{center}
\vskip -0.1in
\end{table*}
We can readily observe that the novel instantiations of BCL  outperform all benchmarks in terms of robustness in Pong, Freeway and BankHeist. The improvement for higher levels of $\epsilon$ is often dramatic. 

For Pong, we observe that both BCL-C-AT-DQN and BCL-MOS-AT-DQN significantly outperform all the benchmark models as well as BCL-RADIAL-DQN for $\epsilon \geq 20/255$, and achieves a near flawless reward. This demonstates the value of our BCL framework as well as the AT curriculum learning approach.

In the Freeway setting, both BCL-MOS-AT-DQN and BCL-RADIAL-DQN achieve high robustness for $\epsilon$ up to $20/255$. In terms of benchmark models, while SA-DQN is competitive at $\epsilon=20/255$, it is far worse at  lower levels of $\epsilon$; for example, when $\epsilon = 10/255$, \Alg-MOS-AT-DQN achieves an average reward that is more than 50\% higher than either SA-DQN or RADIAL-DQN, with DQN (Vanilla) achieving 0 reward at such levels of adversarial perturbations.

Note that we were unable to perform BCL-RADIAL+AT-DQN training for Pong and Freeway. For Pong, this appears to be caused by obfuscated gradients, as we find that the model produced by BCL-RADIAL-DQN have gradients that are nearly zero almost everywhere. For Freeway, since the BCL-RADIAL-DQN result is comparable to BCL-MOS-AT-DQN, BCL-RADIAL-DQN might have already reached the robustness boundary of the AT curriculum training method.

For BankHeist and RoadRunner, BCL-RADIAL+AT-DQN models yield the most significant results. The results demonstrate that performing BCL-C-AT-DQN training on top of BCL-RADIAL-DQN could further improve the model robustness. This again demonstrates the value of our BCL framework as well as our  AT curriculum learning method. Note that for RoadRunner, we find the curriculum training for BCL-RADIAL-DQN is relatively stable, and two of the BCL-RADIAL-DQN  results are achieved by only performing one run for each curriculum phase. Consequently, we include those two runs as the NCL-RADIAL-DQN results.

The AT-DQN and NCL-AT/RADIAL-DQN benchmarks demonstrate the considerable value of the proposed \Alg~framework, which generalizes both methods. In particular, both AT-DQN and NCL-AT-DQN are not competitive for moderate or high values of $\epsilon$, neither with our proposed methods, nor (in most cases) with the other state of the art robust benchmarks.
The NCL-RADIAL-DQN benchmark is more competitive in a few cases (e.g., RoadRunner), but is still significantly outperformed by the best \Alg ~variant.
Note that AT-DQN only performs one training phase with a single fixed $\epsilon$, and we evaluated versions of AT-DQN for many different values of $\epsilon$ values.
The results in Table~\ref{bcl_main_table} for AT-DQN correspond to the best-performing result among these.
In general, as shown in  Appendix \ref{app:dqn_additional}, 
the performance of both AT-DQN and NCL-AT/RADIAL-DQN is relatively unstable.
With AT-DQN in particular, smaller values of $\epsilon$ used in training generally yield poor robustness to stronger attacks, while higher values of $\epsilon$ lead to greater instability and only slightly higher robustness.
In some games, such as Pong, AT-DQN outperforms several other benchmarks (e.g., both SA-DQN and RADIAL-DQN), but it is ineffective in others, such as Freeway.
Similarly, NCL-AT-DQN is also unreliable, working relatively well in some settings (e.g., Freeway), but much worse in others, such as BankHeist.

Our final analysis compares the two proposed methods, \Alg-C-AT-DQN and \Alg-MOS-AT-DQN.
The key advantage of \Alg-MOS-AT-DQN over \Alg-C-AT-DQN is that it potentially significantly reduces training time (in terms of the number of training phases).
And, indeed, it does, as shown in Table~\ref{app:tab_dqn_run_num}: total curriculum training time for all the experiments is reduced by over 50\%, and in the Pong environment, the reduction is over 70\%.

Note that we set the thresholds $\overline{V}(\epsilon)$ quite conservatively, and such thresholds only allow us to skip 1-2 phases each time as observed in the experiments. We can further reduce training time by lowering it, albeit by sacrificing efficacy.
Additionally, we chose an identical threshold for the rewards across all $\epsilon_i$, which typically means that for higher values of $\epsilon$ skipping becomes infrequent as attainable reward drops; making the threshold itself adaptive may further reduce training time.

Surprisingly, however, in addition to the reduction in training time, \Alg-MOS-AT-DQN also typically outperforms \Alg-C-AT-DQN even in efficacy, both in terms of nominal reward and  robustness.
This could be a consequence of opportunistic skipping serving as a form of regularization during training, avoiding overfitting to particular lower-magnitude perturbations.

\begin{table}[!htbp]
\caption{Averaged number of phases of curriculum learning: comparing \Alg-C-AT-DQN and \Alg-MOS-AT-DQN.}
\label{app:tab_dqn_run_num}
\vskip 0.15in
\begin{center}
\begin{small}
\begin{sc}
\begin{tabular}{lllll}
\toprule
            Method/Env.                        & Pong & FW & BH & RR \\\midrule
\Alg-C-AT-DQN   &   66   &     51    &      60     &      36      \\
\Alg-MOS-AT-DQN &   19.3   &    24.0     &      41.7     &     20.3       \\
\bottomrule
\end{tabular}
\end{sc}
\end{small}
\end{center}
\vskip -0.1in
\end{table}

\section{Conclusion}
We purposed a flexible \textit{Bootstrapped Opportunistic Adversarial Curriculum Learning (BCL)} framework. %
The framework allows multiple training runs for each curriculum phase to significantly increase the model stability, as well as opportunistic skipping forward in the curriculum based on custom target reward criteria to improve training efficiency.
We experimentally study four concrete instantiations of the \Alg~framework, varying (a) whether or not we opportunistically skip forward in the curriculum (\Alg-C-AT vs.~\Alg-MOS-AT), and (b) instantiation of the adversarial loss function (\Alg-RADIAL vs.~\Alg-C-AT vs. hybrid \Alg-RADIAL+AT).
In our experiments, \Alg-MOS-AT reduced the training time for all environments by over 50\%  compared to \Alg-C-AT, demonstrating the value of opportunistic skipping.
On the other hand, we find that there is no consistent advantage of one adversarial loss function over the other: in some settings, such as Pong, generating actual adversarial examples leads to far better results, while in others, such as RoadRunner, a combination of both loss functions yields the best performance.
Nevertheless, in all cases the best variant of the proposed \Alg~framework significantly outperforms baselines.

\subsection*{Acknowledgments}

This work was partially supported by the NSF (grants IIS-1905558, IIS-1939677, IIS-1903207, and ECCS-2020289), ARO (grants W911NF1910241 and W911NF1810208), NVIDIA, and Amazon.

\bibliography{robustrl}
\bibliographystyle{icml2022}

\newpage
\appendix
\onecolumn
\section{Bootstrapped Opportunistic Adversarial Curriculum Learning (BCL) for PPO}
\label{app:ppo_main_results}
In this section, we extend the application of BCL framework to PPO-style approaches. We evaluate our approaches on two Procgen~\cite{cobbe2020leveraging} environments: FruitBot and Jumper. For FruitBot, both AT-PPO and BCL-MOS-AT-PPO show higher nominal rewards and significant improvements in terms of robustness for up to 20/255, while the current existing vanilla PPO and RADIAL-PPO has small or even negative rewards  for $\epsilon \geq 10/255$. For Jumper, BCL-MOS-AT-PPO  achieves significant improvements in terms of robustness for up to 40/255, with rewards under adversarial attacks for $\epsilon \geq 10/255$ more than doubled compare to vanilla PPO (the current most robust model).  The experiments on Procgen also demonstrate that our models exhibit good generalization, as the evaluation rewards are high under both training and evaluation distributions.

\paragraph{PPO} PPO~\cite{schulman2017proximal} is a policy gradient method for reinforcement learning, with the objective function as
\begin{align}
\label{standard_ppo_loss}
    \mathcal{L}(\theta)=\mathbb{E}_{\left(s_{t}, a_{t}, r_{t}\right)}\left[-\min \left(\frac{\pi\left(a_{t} | s_{t} ; \theta\right)}{\pi\left(a_{t} | s_{t} ; \theta_{\text {old}}\right)} A_{t}, \operatorname{clip}\left(\frac{\pi\left(a_{t} | s_{t} ; \theta\right)}{\pi\left(a_{t} | s_{t} ; \theta_{\text {old}}\right)}, 1-\eta, 1+\eta\right) A_{t}\right)\right].
\end{align}
Here $\pi$ is the policy, $A_t$ is the advantage function at time $t$ and  $\eta$ is the hyperparameter. PPO modifies the surrogate objective by clipping the policy ratio to constrain the difference between old and new policy, which stabilizes the training and speeds up convergence. We use  Equation \eqref{standard_ppo_loss} as the $\mathcal{L}_{\text{standard}}$ in Equation \eqref{eq:radial_loss}.

\paragraph{RADIAL-PPO}RADIAL-PPO \cite{Oikarinen2021} defines the  adversarial loss function as
 \begin{align*}
      \mathcal{L}_{\text{adv}}(\theta, \epsilon)=\mathbb{E}_{\left(s_{t}, a_{t}, r_{t}\right)}
      \left[-\min \left(\frac{{\pi}^\epsilon\left(a_{t} | s_{t}, \epsilon ; \theta\right)}{\pi\left(a_{t} | s_{t} ; \theta_{\text {old}}\right)} A_{t},\operatorname{clip}\left(\frac{{\pi}^\epsilon\left(a_{t} | s_{t}, \epsilon ; \theta\right)}{\pi\left(a_{t} | s_{t} ; \theta_{\text {old}}\right)}, 1-\eta, 1+\eta\right) A_{t}\right)\right],
 \end{align*}
with ${\pi}^\epsilon$ the lower bound of the policy network if $A_t \geq 0$, and upper bound otherwise. 

The goal for $\mathcal{L}_{\text{adv}}$ in RADIAL-PPO is to form a strict upper bound of the loss function under adversarial perturbations, that is, $\mathcal{L}_{\text{standard}}(s+\delta;\epsilon) \leq \mathcal{L}_{\text{adv}}(s; \epsilon)$ with $||\delta||_p \leq \epsilon$. Robustness is achieved through constraining the strict upper bound of the loss function. This is referred to as approach \# 1 in \citet{Oikarinen2021}, and was used for RADIAL-PPO training.  %

\paragraph{Generating Adversarial Perturbations}In our model, we replace the upper and lower bounds $\pi^\epsilon(a|s_t)$  in RADIAL-PPO with $\tilde{\pi}_1^\epsilon(a + \delta^*|s_t)$ and $\tilde{\pi}_2^\epsilon(a + \delta^*|s_t)$. For RI-FGSM, $\delta^*$ (approximately) solves the following optimization problem:
\begin{equation}
    \min_{||\delta||_\infty \leq \epsilon}
    \mathrm{Softmax} (\mathrm{Logits}(s + \delta))\odot \mathrm{Logits}(s + \delta),
    \label{eq:PPO_pert}
\end{equation}
where $\mathrm{Logits}(s)$  is the output vector of the PPO-style neural network and is used to calculate the categorical distribution $\pi(s)$. Note that $\mathrm{Logits}(s)$ has the same size as the action space. For PGD attacks, it is to (approximately) maximize  ${\mathcal{L}(\mathrm{Softmax}(\mathrm{Logits}(s+\delta^*;\theta)),\pi(s))}$ with respect to $\delta^*$, where $\mathcal{L}$ is the cross-entropy loss and $\delta^*$ is updated iteratively over a fixed number of iterations (same as in the PGD attack for DQN models).

Since $\pi$ is a categorical distribution over possible (discrete) actions, we calculate $\tilde{\pi}_1^\epsilon(a + \delta^*|s_t)$  by having the $a$-th logit as the one under adversarial perturbation, and the rest are  vanilla logits; $\tilde{\pi}_2^\epsilon(a + \delta^*|s_t)$ is calculated by having the $a$-th logit as the vanilla logit, and the rest are logits under adversarial perturbation. The adversarial loss function is defined as the maximum loss under those two policies:
 \begin{align}
      \mathcal{L}_{\text{adv}}(\theta, \epsilon)=\mathbb{E}_{\left(s_{t}, a_{t}, r_{t}\right)}
      \left[-\min_{i \in \{1,2\}} \min\left(\frac{\tilde{\pi}_i^\epsilon\left(a_{t} | s_{t}, \epsilon ; \theta\right)}{\pi\left(a_{t} | s_{t} ; \theta_{\text {old}}\right)} A_{t},\operatorname{clip}\left(\frac{\tilde{\pi}_i^\epsilon\left(a_{t} | s_{t}, \epsilon ; \theta\right)}{\pi\left(a_{t} | s_{t} ; \theta_{\text {old}}\right)}, 1-\eta, 1+\eta\right) A_{t}, \right)\right].
      \label{ppo_adv_func}
 \end{align}

Note that instead of forming a strict upper bound of the adversarial loss function as in RADIAL-PPO, $\mathcal{L}_{\text{adv}}$ in Equation \eqref{ppo_adv_func} provides a lower bound of the adversarial loss function with heuristic adversarial examples. %

\paragraph{Experiment Setup} We evaluate the purposed approach using two Procgen environments~\cite{cobbe2020leveraging} with discrete action space: FruitBot and Jumper. Note that we did not experiment on the CoinRun environment as in \citet{Oikarinen2021}. We find that for CoinRun environment there is an optimal action: we could achieve a reward comparable to  RADIAL-PPO regardless of the magnitude of $\epsilon$.  For each model, we calculate an efficacy score $\sum_{\text{dist}\in\{\text{Train, Eval}\} }\left(R_{\text{nominal}}^{\text{dist}} + \frac{1}{3}\sum_\epsilon R_{\text{PGD}}^{ \text{dist}, \epsilon}\right)$ for all $\epsilon$ listed in Table \ref{bcl_ppo_main}, where $R_{\text{nominal}}^{\text{dist}}$ is the nominal reward and $R_{\text{PGD}}^{ \text{dist}, \epsilon}$ is the reward under 30-step PGD attack with adversarial perturbation size $||\delta||_\infty \leq \epsilon$ under Train/Eval distribution. Similar to DQN experiments, we conduct three independent runs for each experiment based on the efficacy score and present the median result in the main table (Table \ref{bcl_ppo_main}). For AT-PPO, we conduct experiments for all  the three $\epsilon$ listed in Table \ref{bcl_ppo_main}, select the median run for each AT-PPO-$\epsilon$, and present the best result  as the strongest benchmark.%

We find that the adversarial training with PPO-style approaches is relatively stable, and that $K=1$ suffices.  Thus, we use PPO (Vanilla), RADIAL-PPO \cite{Oikarinen2021} as well as AT-PPO as benchmarks. For PPO (Vanilla) and RADIAL-PPO we use the results from \citet{Oikarinen2021}, and we perform our own AT-PPO training with the method purposed above as a restricted version of BCL. We  did not perform curriculum learning with RADIAL as it does not work for Jumper.
   For FruitBot, we use RI-FGSM to generate adversarial examples. For Jumper, as we find RI-FGSM is not effective, we instead use 10-step PGD to generate adversarial examples.
 We skipped BCL-C-AT-PPO experiments due to extensive computational costs; however, as we will show in the results section, the opportunistic skipping forward mechanism under the BCL framework makes the training possible for Jumper even with 10-step PGD. 
 
 To ensure a fair comparison, we let all methods to have the same computational constraints and evaluation metrics: for all environments we train for 25 million steps on the easy setting for each run. For evaluation, we use 30-step PGD attack with step size 0.1 for all models, which is stronger than the 10-step PGD attack used in \citet{Oikarinen2021}. We evaluate all models over 1000 episodes using deterministic policy and report the averaged reward under both training distribution (easy setting) and evaluation distribution (full distribution), which is the same as in \citet{Oikarinen2021}.

\paragraph{Hyperparameters} The PPO specific hyperparameters as well as $\kappa$ for AT-PPO and BCL-MOS-AT-PPO  are the same as in RADIAL-PPO. In FruitBot we use RI-FGSM to generate adversarial examples, with hyperparameter $\alpha = 95.5$, which is approximately $0.375 \times 255$ (note that $\alpha = 0.375$ is used in DQN experiments for RI-FGSM). This is due to in RADIAL-PPO code when the gradients are calculated the state space has a range of $0\sim 255$, instead of being normalized to $0\sim 1$ as in RADIAL-DQN.  The thresholds for BCL-MOS-AT-PPO are shown in Table \ref{TH_RADIAL_PPO}, where $\overline{V}_{\text{PGD(Train)}}(\epsilon)$ is the threshold for the averaged reward under 30-step PGD  attack under training distribution.

\begin{table}[htbp!]
\caption{Thresholds $\overline{V}(\epsilon)$ for BCL-MOS-AT-PPO}
\label{TH_RADIAL_PPO}
\vskip 0.15in
\begin{center}
\begin{small}
\begin{sc}
\begin{tabular}{lll}
\toprule
  Criteria/Env.        & FruitBot & Jumper   \\\midrule
  $\overline{V}_{\text{PGD(Train)}}(\epsilon)$   & 25.0   & 6.0     \\
\bottomrule
\end{tabular}
\end{sc}
\end{small}
\end{center}
\vskip -0.1in
\end{table}

We find that although RADIAL-PPO increases the robustness for lower $\epsilon$ compared to vanilla PPO (e.g., $\epsilon = 5/255$ under 10-step PGD attack as shown in \citet{Oikarinen2021}), it decreases the robustness for higher $\epsilon$, accompanied by  a lower nominal reward. Thus we perform two sets of BCL-MOS-AT-PPO experiments: 1) BCL-MOS(V)-AT-PPO, where we bootstrap from PPO (Vanilla); and 2) BCL-MOS(R)-AT-PPO, where  we bootstrap from RADIAL-PPO.

Our baseline curriculum for PPO has an increment of 1/255, with $\epsilon_0$ for each experiment shown in Table \ref{baseline_cl_PPO}. For FruitBot we set target $\epsilon = 20/255$, and for Jumper we set target $\epsilon = 40/255$. For all experiments we set $K=1$.  We stop the curriculum training when the model is robust against the target $\epsilon$, meaning the reward under training distribution is above the threshold in Table \ref{TH_RADIAL_PPO}. Note that for FruitBot, we also stop the training when RI-FGSM attack is ineffective towards the  target $\epsilon$, meaning that although 30-step PGD indicates the model is not robust against $\epsilon_i$, however, RI-FGSM returns a near perfect reward (close to nominal reward) when the magnitude of adversarial perturbation for the attack is  $\epsilon$. In this case, since RI-FGSM is not generating any meaningful adversarial examples, continue training will in fact decrease the model robustness. %

\begin{table}[htbp!]
\caption{BCL-MOS-AT-PPO specific hyperparameters ($\epsilon_0$).}
\label{baseline_cl_PPO}
\vskip 0.15in
\begin{center}
\begin{small}
\begin{sc}
\begin{tabular}{lll}
\toprule
  Criteria/Env.        & FruitBot & Jumper   \\\midrule
    BCL-MOS(V)-AT-PPO   & 0/255   &   1/255 \\
  BCL-MOS(R)-AT-PPO   & 6/255   &  9/255    \\
\bottomrule
\end{tabular}
\end{sc}
\end{small}
\end{center}
\vskip -0.1in
\end{table}

\paragraph{Walltime} Each adversarial training run  takes 10 hours for FruitBot (with RI-FGSM), and 34 hours for Jumper (with 10-step PGD) on a single GeForce RTX 2080Ti GPU. The number of runs conducted for each experiment is shown in Table \ref{ppo_walltime}.

\begin{table}[htbp!]
\caption{Number of runs conducted for each experiment.}%
\label{ppo_walltime}
\vskip 0.15in
\begin{center}
\begin{small}
\begin{sc}
\begin{tabular}{llll}
\toprule\toprule
\multicolumn{4}{c}{FruitBot}               \\
                  & Run1 & Run2 & Run3 \\\midrule
AT-PPO-$\epsilon$        & 1    & 1    & 1    \\
BCL-MOS(V)-AT-PPO    & 3   & 2   & 2   \\
BCL-MOS(R)-AT-PPO    &  2  &  2  &  2  \\
\bottomrule\toprule
\multicolumn{4}{c}{Jumper}            \\
                  & Run1 & Run2 & Run3 \\\midrule
AT-PPO-$\epsilon$        & 1    & 1    & 1    \\
BCL-MOS(V)-AT-PPO    &  4  &    4& 4   \\
BCL-MOS(R)-AT-PPO    &    4&   4 &4    \\
\bottomrule
\end{tabular}

\end{sc}
\end{small}
\end{center}
\vskip -0.1in
\end{table}

\paragraph{Results} Our main results are shown in Table \ref{bcl_ppo_main}, with detailed results deferred to Table \ref{ppo_fruit_detail} and Table \ref{ppo_jumper_detail}.  The results show that the models trained with our BCL-MOS-AT-PPO approach exhibit significant improvements in terms of nominal reward, robustness as well as generalization compared to state of the art  PPO (Vanilla) and RADIAL-PPO.

For FruitBot, both AT-PPO and BCL-MOS-AT-PPO achieve significant improvements in terms of robustness for $\epsilon$ up to 20/255, while  PPO (Vanilla) and RADIAL-PPO has small or even negative rewards for $\epsilon \geq 10/255$.  Furthermore,  our AT models achieve higher  nominal rewards  under both training and evaluation distributions. The success of AT-PPO and BCL-MOS-AT-PPO demonstrate the value of our approach for training with adversarial examples.

For Jumper,  BCL-MOS(V)-AT-PPO achieves the most robust model, with rewards under both training and evaluation distributions significantly outperform all benchmark models for $\epsilon \geq 10/255$; it also has a higher nominal reward under evaluation distribution, and  comparable high nominal reward under training distribution. Furthermore, we find that BCL-MOS(V)-AT-PPO outperforms BCL-MOS(R)-AT-PPO both in terms of nominal rewards and robustness for  $\epsilon \geq 20/255$. This is mainly because RADIAL-PPO  has a lower nominal reward and is less robust compared to vanilla PPO  for $\epsilon \geq 20/255$. 

We also find that with BCL-MOS-AT-PPO training, the resulting models can be easily robust against a higher $\epsilon$ when trained against a lower one. The final models for BCL-MOS-AT-PPO in the Jumper environment are only trained for 4 curriculum phases to achieve robustness against $\epsilon$ = 40/255. This demonstrates the value of the opportunistic skipping forward mechanism in our BCL framework.

\begin{table}[htbp!]
\caption{Average episode rewards $\pm$ standard error of the mean (SEM) over 1000 episodes on both training and evaluation set. The \colorbox{Gray}{\textcolor{red}{gray rows}} are the most robust models, selected based on efficacy score $\sum_{\text{dist} }\left(R_{\text{nominal}}^{\text{dist}} + \frac{1}{3}\sum_\epsilon R_{\text{PGD}}^{ \text{dist}, \epsilon}\right)$. \textbf{Boldface} marks the best results for each value of $\epsilon$; we marked multiple row entries as boldface for a given $\epsilon$ if they are statistically indistinguishable (i.e., have overlapping confidence intervals).}
\label{bcl_ppo_main}
\vskip 0.15in
\begin{center}
\begin{small}
\begin{sc}
\begin{tabular}{llllll}
\toprule\toprule
\multicolumn{6}{c}{\textbf{FruitBot}}                                                                           \\ 
 Model                & Dist. & Nominal      &\multicolumn{3}{c}{30-step PGD Attack}  \\\midrule
 & &  $\epsilon$ = 0  & $\epsilon$ = 10/255 &$\epsilon$= 15/255 &$\epsilon$= 20/255\\\toprule
PPO (Vanilla)                   & Train &$30.20 \pm 0.23 $ & $ 2.40 \pm 0.21  $ & $ 0.73 \pm 0.16  $ & $ -0.72 \pm 0.14$ \\
                       & Eval   &${26.09 \pm 0.33} $ & $ 1.70 \pm 0.20  $ & $ 0.11 \pm 0.14  $ & $ -0.50 \pm 0.13$ \\\cmidrule{1-1}
RADIAL-PPO             & Train  &$28.03 \pm 0.24 $ & $ -0.90 \pm 0.13 $ & $ -1.28 \pm 0.10 $ & $ -1.64 \pm 0.10$ \\
                       & Eval   &${26.08 \pm 0.29 }$ & $ -1.24 \pm 0.13 $ & $ -1.53 \pm 0.11 $ & $ -1.81 \pm 0.11$\\
                       \midrule
{AT-PPO}          & Train &${31.14 \pm 0.19} $ & $ {28.69 \pm 0.29 }$ & ${ 26.35 \pm 0.32 }$ & $\mathbf{ 24.41 \pm 0.35} $\\
                        & Eval  &$\mathbf{28.26 \pm 0.29 }$ & ${ 26.47 \pm 0.34 }$ & $ {24.56 \pm 0.36} $ & $ {20.44 \pm 0.40 }$\\
                       \midrule
\rowcolor{Gray}\textcolor{red}{BCL-MOS(V)-AT-PPO} & Train  &$\mathbf{32.11 \pm 0.17} $ & $ 29.98 \pm 0.24 $ & $ 27.40 \pm 0.31 $ & $ \mathbf{24.23 \pm 0.36}$\\
\rowcolor{Gray}                      & Eval    &$\mathbf{28.81 \pm 0.28} $ & $\mathbf{ 27.61 \pm 0.31} $ & $\mathbf{ 25.52 \pm 0.35 }$ & $ \mathbf{21.63 \pm 0.39}$\\\cmidrule{1-1}
{BCL-MOS(R)-AT-PPO} & Train  &$31.40 \pm 0.20 $ & $ \mathbf{30.80 \pm 0.21} $ & $ \mathbf{28.22 \pm 0.30} $ & $ 20.18 \pm 0.40$ \\
                      & Eval    & $26.95 \pm 0.34 $ & $ 26.28 \pm 0.35 $ & $ 24.17 \pm 0.37 $ & $ 17.87 \pm 0.41$\\
 \bottomrule
\toprule
\multicolumn{6}{c}{\textbf{Jumper}}                                                                           \\ 
 Model                & Dist. & Nominal      &\multicolumn{3}{c}{30-step PGD Attack}  \\\midrule
 & &  $\epsilon$ = 0  & $\epsilon$ = 10/255 &$\epsilon$= 20/255 &$\epsilon$= 40/255\\\toprule
PPO (Vanilla)                    & Train &$\mathbf{8.69 \pm 0.11} $ & $ 3.42 \pm 0.15 $ & $ 3.61 \pm 0.15 $ & $2.94 \pm 0.14$\\
                       & Eval  &$4.22 \pm 0.16 $ & $ 2.81 \pm 0.14 $ & $ 2.62 \pm 0.14 $ &$2.50 \pm 0.14$ \\\cmidrule{1-1}
RADIAL-PPO             & Train &$6.59 \pm 0.15 $ & $ 5.43 \pm 0.16 $ & $ 2.45 \pm 0.14 $ & $1.44 \pm 0.11$\\
                       & Eval  &$3.85 \pm 0.15 $ & $ 3.03 \pm 0.14 $ & $ 2.04 \pm 0.13 $ &$1.44 \pm 0.11$ \\
                       \midrule
AT-PPO          & Train &$7.57 \pm 0.14$ & $4.98 \pm 0.16$ & $4.35 \pm 0.16$ & $3.52 \pm 0.15$ \\
                       & Eval  &$\mathbf{4.55 \pm 0.16}$ & $3.81 \pm 0.15$ & $3.35 \pm 0.15$ & $2.51 \pm 0.14$ \\
                       \midrule
\rowcolor{Gray}\textcolor{red}{BCL-MOS(V)-AT-PPO}  & Train &$ \mathbf{8.67 \pm 0.11} $ & $ \mathbf{8.15 \pm 0.12} $ & $ \mathbf{8.40 \pm 0.12} $ & $\mathbf{ 7.84 \pm 0.13 }$\\
\rowcolor{Gray}                       & Eval &$ \mathbf{4.57 \pm 0.16} $ & $\mathbf{ 4.64 \pm 0.16} $ & $\mathbf{ 4.65 \pm 0.16 }$ & $ \mathbf{4.41 \pm 0.16} $\\\cmidrule{1-1}
{BCL-MOS(R)-AT-PPO}  & Train &$ 8.09 \pm 0.12 $ & $ \mathbf{8.29 \pm 0.12 }$ & $ \mathbf{8.40 \pm 0.12} $ & $ 6.93 \pm 0.15 $ \\
              &        Eval &$\mathbf{ 4.39 \pm 0.16} $ & $ 4.29 \pm 0.16 $ & $ 4.09 \pm 0.16 $ & $ 3.85 \pm 0.15 $ \\

 \bottomrule
\end{tabular}
\end{sc}
\end{small}
\end{center}
\vskip -0.1in
\end{table}

\section{Additional Experimental Results for PPO models}
\label{app:ppo_additional}
Below we show the detailed experimental results for all three independent runs for each experiment. We separate the results by environments for better visualization and comparison. For PPO (Vanilla) and RADIAL-PPO we use the released models from \citet{Oikarinen2021}. For AT-PPO, we perform our own training as a restricted case of the BCL algorithm. 

For AT-PPO, we include all 9 runs for each environment, naming them as AT-PPO-$\epsilon$, where $\epsilon$ is the single value of the base curriculum taken as an input by the model. The choices of $\epsilon$ are the ones we use for evaluation. For each AT-PPO-$\epsilon$ we conduct three independent runs, and use the median of those three runs as the representative result. The result reported in Table \ref{bcl_ppo_main}  for each environment is the best median result among those three $\epsilon$.

\begin{table}[htbp!]
\caption{FruitBot environment. Average episode rewards $\pm$ standard error of the mean (SEM) over 1000 episodes on both training and evaluation set. The \colorbox{Gray}{{gray rows}} are the median of three runs, selected based on efficacy score $\sum_{\text{dist} }\left(R_{\text{nominal}}^{\text{dist}} + \frac{1}{3}\sum_\epsilon R_{\text{PGD}}^{ \text{dist}, \epsilon}\right)$.}
\label{ppo_fruit_detail}
\vskip 0.15in
\begin{center}
\begin{small}
\begin{sc}
\begin{tabular}{llllll}
\toprule\toprule
\multicolumn{6}{c}{\textbf{FruitBot}}                                                                           \\ 
 Model                & Dist. & Nominal      &\multicolumn{3}{c}{30-step PGD Attack}  \\\midrule
 & &  $\epsilon$ = 0  & $\epsilon$ = 10/255 &$\epsilon$= 15/255 &$\epsilon$= 20/255\\\toprule
PPO (Vanilla)                   & Train &$30.20 \pm 0.23 $ & $ 2.40 \pm 0.21  $ & $ 0.73 \pm 0.16  $ & $ -0.72 \pm 0.14$ \\
                       & Eval  &$26.09 \pm 0.33 $ & $ 1.70 \pm 0.20  $ & $ 0.11 \pm 0.14  $ & $ -0.50 \pm 0.13$ \\\cmidrule{1-1}
RADIAL-PPO             & Train &$28.03 \pm 0.24 $ & $ -0.90 \pm 0.13 $ & $ -1.28 \pm 0.10 $ & $ -1.64 \pm 0.10$ \\
                       & Eval  &$26.08 \pm 0.29 $ & $ -1.24 \pm 0.13 $ & $ -1.53 \pm 0.11 $ & $ -1.81 \pm 0.11$\\
                       \midrule
AT-PPO-10/255 (Run1)          & Train &$24.69 \pm 0.46 $ & $ 23.46 \pm 0.46 $ & $ 22.38 \pm 0.45 $ & $ 17.73 \pm 0.47$ \\
                       & Eval  &$24.38 \pm 0.43 $ & $ 23.69 \pm 0.42 $ & $ 21.66 \pm 0.45 $ & $ 17.26 \pm 0.46$ \\
                       \cmidrule{1-1}
\rowcolor{Gray}AT-PPO-10/255 (Run2)           & Train &$30.27 \pm 0.23 $ & $ 28.73 \pm 0.29 $ & $ 27.03 \pm 0.31 $ & $ 22.97 \pm 0.36$ \\
\rowcolor{Gray}                       & Eval  &$28.15 \pm 0.28 $ & $ 27.30 \pm 0.30 $ & $ 25.17 \pm 0.34 $ & $ 20.42 \pm 0.40$ \\
                       \cmidrule{1-1}
AT-PPO-10/255 (Run3)            & Train &$31.20 \pm 0.19 $ & $ 31.08 \pm 0.17 $ & $ 29.13 \pm 0.23 $ & $ 23.83 \pm 0.36$ \\
                       & Eval  &$28.93 \pm 0.27 $ & $ 27.72 \pm 0.30 $ & $ 25.99 \pm 0.33 $ & $ 21.91 \pm 0.39$\\
                       \midrule                   
AT-PPO-15/255 (Run1)             & Train  &$29.62 \pm 0.27 $ & $ 28.92 \pm 0.29 $ & $ 26.82 \pm 0.32 $ & $ 22.47 \pm 0.38$ \\
                       & Eval   &$26.48 \pm 0.34 $ & $ 25.71 \pm 0.35 $ & $ 24.10 \pm 0.37 $ & $ 22.00 \pm 0.38$ \\
                       \cmidrule{1-1}
AT-PPO-15/255 (Run2)            & Train  &$31.48 \pm 0.18 $ & $ 29.49 \pm 0.26 $ & $ 28.82 \pm 0.27 $ & $ 24.72 \pm 0.35$ \\
                       & Eval   &$28.48 \pm 0.28 $ & $ 27.30 \pm 0.30 $ & $ 25.97 \pm 0.32 $ & $ 22.82 \pm 0.37$\\
                       \cmidrule{1-1}
\rowcolor{Gray}AT-PPO-15/255 (Run3)             & Train  &$31.04 \pm 0.23 $ & $ 28.75 \pm 0.29 $ & $ 26.76 \pm 0.34 $ & $ 20.92 \pm 0.42$ \\
\rowcolor{Gray}                      & Eval   &$28.07 \pm 0.31 $ & $ 26.41 \pm 0.34 $ & $ 24.09 \pm 0.38 $ & $ 20.33 \pm 0.41$ \\
                       \midrule                 
AT-PPO-20/255 (Run1)             & Train &$30.65 \pm 0.23 $ & $ 27.83 \pm 0.32 $ & $ 26.16 \pm 0.34 $ & $ 21.04 \pm 0.39 $\\
                       & Eval   &$27.44 \pm 0.32 $ & $ 25.64 \pm 0.35 $ & $ 22.67 \pm 0.39 $ & $ 20.57 \pm 0.40 $\\
                       \cmidrule{1-1}   
\rowcolor{Gray}AT-PPO-20/255 (Run2)            & Train  &$31.14 \pm 0.19 $ & $ 28.69 \pm 0.29 $ & $ 26.35 \pm 0.32 $ & $ 24.41 \pm 0.35 $\\
\rowcolor{Gray}                       & Eval   &$28.26 \pm 0.29 $ & $ 26.47 \pm 0.34 $ & $ 24.56 \pm 0.36 $ & $ 20.44 \pm 0.40 $\\
                       \cmidrule{1-1}   
AT-PPO-20/255 (Run3)             & Train  &$28.62 \pm 0.31 $ & $ 29.83 \pm 0.24 $ & $ 28.12 \pm 0.27 $ & $ 25.96 \pm 0.32 $\\
                       & Eval   &$27.90 \pm 0.30 $ & $ 27.20 \pm 0.31 $ & $ 25.62 \pm 0.34 $ & $ 23.55 \pm 0.35 $\\        
                       \midrule
\midrule
BCL-MOS(V)-AT-PPO (Run1) & Train &$31.32 \pm 0.21 $ & $ 30.34 \pm 0.23 $ & $ 28.64 \pm 0.30 $ & $ 26.24 \pm 0.35$ \\
& Eval &$28.94 \pm 0.27 $ & $ 27.57 \pm 0.30 $ & $ 26.34 \pm 0.32 $ & $ 23.55 \pm 0.36$  \\\cmidrule{1-1}
BCL-MOS(V)-AT-PPO (Run2) & Train &$31.37 \pm 0.22 $ & $ 29.79 \pm 0.26 $ & $ 26.64 \pm 0.33 $ & $ 24.38 \pm 0.35$ \\
& Eval & $28.06 \pm 0.31 $ & $ 26.51 \pm 0.33 $ & $ 24.41 \pm 0.36 $ & $ 22.12 \pm 0.39$  \\\cmidrule{1-1}
\rowcolor{Gray}BCL-MOS(V)-AT-PPO (Run3) & Train &$32.11 \pm 0.17 $ & $ 29.98 \pm 0.24 $ & $ 27.40 \pm 0.31 $ & $ 24.23 \pm 0.36$ \\
\rowcolor{Gray}& Eval &$28.81 \pm 0.28 $ & $ 27.61 \pm 0.31 $ & $ 25.52 \pm 0.35 $ & $ 21.63 \pm 0.39$\\\midrule
BCL-MOS(R)-AT-PPO (Run1) & Train & $31.13 \pm 0.22 $ & $ 28.91 \pm 0.26 $ & $ 26.56 \pm 0.32 $ & $ 22.37 \pm 0.37 $\\
                      & Eval  & $ 26.62 \pm 0.34 $ & $ 24.67 \pm 0.37 $ & $ 21.76 \pm 0.39 $ & $ 19.54 \pm 0.41 $\\\cmidrule{1-1}
BCL-MOS(R)-AT-PPO (Run2) & Train & $30.85 \pm 0.21 $ & $ 29.77 \pm 0.25 $ & $ 26.93 \pm 0.34 $ & $ 23.03 \pm 0.37$ \\
                      & Eval  & $27.30 \pm 0.32 $ & $ 25.98 \pm 0.34 $ & $ 24.54 \pm 0.35 $ & $ 21.14 \pm 0.39$ \\\cmidrule{1-1}
\rowcolor{Gray}BCL-MOS(R)-AT-PPO (Run3) & Train&$31.40 \pm 0.20 $ & $ 30.80 \pm 0.21 $ & $ 28.22 \pm 0.30 $ & $ 20.18 \pm 0.40$ \\
\rowcolor{Gray}                      & Eval & $26.95 \pm 0.34 $ & $ 26.28 \pm 0.35 $ & $ 24.17 \pm 0.37 $ & $ 17.87 \pm 0.41$\\
\bottomrule
\end{tabular}
\end{sc}
\end{small}
\end{center}
\vskip -0.1in
\end{table}

\begin{table}[htbp!]
\caption{Jumper environment. Average episode rewards $\pm$ standard error of the mean (SEM) over 1000 episodes on both training and evaluation set. The \colorbox{Gray}{{gray rows}} are the median of three runs, selected based on efficacy score $\sum_{\text{dist} }\left(R_{\text{nominal}}^{\text{dist}} + \frac{1}{3}\sum_\epsilon R_{\text{PGD}}^{ \text{dist}, \epsilon}\right)$.}
\label{ppo_jumper_detail}
\vskip 0.15in
\begin{center}
\begin{small}
\begin{sc}
\begin{tabular}{llllll}
\toprule\toprule
\multicolumn{6}{c}{\textbf{Jumper}}                                                                           \\ 
 Model                & Dist. & Nominal      &\multicolumn{3}{c}{30-step PGD Attack}  \\\midrule
 & &  $\epsilon$ = 0  & $\epsilon$ = 10/255 &$\epsilon$= 20/255 &$\epsilon$= 40/255\\\toprule
PPO (Vanilla)                    & Train &$8.69 \pm 0.11 $ & $ 3.42 \pm 0.15 $ & $ 3.61 \pm 0.15 $ & $2.94 \pm 0.14$\\
                       & Eval  &$4.22 \pm 0.16 $ & $ 2.81 \pm 0.14 $ & $ 2.62 \pm 0.14 $ &$2.50 \pm 0.14$ \\\cmidrule{1-1}
RADIAL-PPO             & Train &$6.59 \pm 0.15 $ & $ 5.43 \pm 0.16 $ & $ 2.45 \pm 0.14 $ & $1.44 \pm 0.11$\\
                       & Eval  &$3.85 \pm 0.15 $ & $ 3.03 \pm 0.14 $ & $ 2.04 \pm 0.13 $ & $1.44 \pm 0.11$\\
                       \midrule
AT-PPO-10/255 (Run1)          & Train &$6.58 \pm 0.15 $ & $ 6.96 \pm 0.15 $ & $ 5.54 \pm 0.16 $ & $0.97 \pm 0.09$\\
                       & Eval  &$4.39 \pm 0.16 $ & $ 4.24 \pm 0.16 $ & $ 3.30 \pm 0.15 $ & $0.43 \pm 0.06$\\
                       \cmidrule{1-1}
\rowcolor{Gray}AT-PPO-10/255 (Run2)           & Train &$7.41 \pm 0.14 $ & $ 6.70 \pm 0.15 $ & $ 5.04 \pm 0.16 $ & $1.31 \pm 0.11$\\
\rowcolor{Gray}                      & Eval  &$4.50 \pm 0.16 $ & $ 4.05 \pm 0.16 $ & $ 3.35 \pm 0.15 $ & $0.68 \pm 0.08$\\
                       \cmidrule{1-1}
AT-PPO-10/255 (Run3)            & Train &$7.47 \pm 0.14 $ & $ 6.94 \pm 0.15 $ & $ 5.29 \pm 0.16 $ & $1.02 \pm 0.10$\\
                       & Eval  &$4.64 \pm 0.16 $ & $ 4.31 \pm 0.16 $ & $ 3.27 \pm 0.15 $ & $0.58  \pm 0.07$\\
                       \midrule                   
AT-PPO-20/255 (Run1)             & Train  &$6.90 \pm 0.15 $ & $ 6.35 \pm 0.15 $ & $ 5.23 \pm 0.16 $ & $2.41 \pm 0.14$\\
                       & Eval   &$4.47 \pm 0.16 $ & $ 4.31 \pm 0.16 $ & $ 3.92 \pm 0.15 $ & $1.46 \pm 0.11$\\
                       \cmidrule{1-1}
AT-PPO-20/255 (Run2)            & Train  &$5.96 \pm 0.16 $ & $ 5.98 \pm 0.16 $ & $ 5.05 \pm 0.16 $ & $1.83 \pm 0.12$\\
                       & Eval   &$4.56 \pm 0.16 $ & $ 4.29 \pm 0.16 $ & $ 3.85 \pm 0.15 $ & $1.78 \pm 0.12$\\
                       \cmidrule{1-1}
\rowcolor{Gray}AT-PPO-20/255 (Run3)             & Train  &$6.56 \pm 0.15 $ & $ 6.14 \pm 0.15 $ & $ 5.06 \pm 0.16 $ & $1.85 \pm 0.12$\\
\rowcolor{Gray}                      & Eval   &$4.42 \pm 0.16 $ & $ 4.34 \pm 0.16 $ & $ 4.00 \pm 0.15 $ & $1.27 \pm 0.11$\\
                       \midrule                 
\rowcolor{Gray}AT-PPO-40/255 (Run1)       &Train      &$7.57 \pm 0.14$ & $4.98 \pm 0.16$ & $4.35 \pm 0.16$ & $3.52 \pm 0.15$ \\
\rowcolor{Gray}                       & Eval  &$4.55 \pm 0.16$ & $3.81 \pm 0.15$ & $3.35 \pm 0.15$ & $2.51 \pm 0.14$ \\
                       \cmidrule{1-1}   
AT-PPO-40/255 (Run2)            & Train  &$7.43 \pm 0.14$ & $4.74 \pm 0.16$ & $4.20 \pm 0.16$ & $3.98 \pm 0.15$ \\
                     & Eval   &$4.54 \pm 0.16$ & $3.88 \pm 0.15$ & $3.24 \pm 0.15$ & $3.39 \pm 0.15$ \\
                       \cmidrule{1-1}   
AT-PPO-40/255 (Run3)             & Train  &$6.72 \pm 0.15$ & $4.66 \pm 0.16$ & $4.36 \pm 0.16$ & $4.01 \pm 0.16$ \\
                       & Eval   &$4.71 \pm 0.16$ & $3.90	\pm 0.15$ & $3.19 \pm 0.15$ & $2.76 \pm 0.14$ \\                
                       \midrule\midrule
\rowcolor{Gray}BCL-MOS(V)-AT-PPO (Run1) &Train &$ 8.67 \pm 0.11 $ & $ 8.15 \pm 0.12 $ & $ 8.40 \pm 0.12 $ & $ 7.84 \pm 0.13 $\\
\rowcolor{Gray}& Eval &$ 4.57 \pm 0.16 $ & $ 4.64 \pm 0.16 $ & $ 4.65 \pm 0.16 $ & $ 4.41 \pm 0.16 $\\\cmidrule{1-1}   
BCL-MOS(V)-AT-PPO (Run2) &Train &$ 9.09 \pm 0.09 $ & $ 8.85 \pm 0.10 $ & $ 8.50 \pm 0.11 $ & $ 7.64 \pm 0.13 $\\
& Eval &$ 4.77 \pm 0.16 $ & $ 4.77 \pm 0.16 $ & $ 4.78 \pm 0.16 $ & $ 4.43 \pm 0.16 $ \\\cmidrule{1-1}   
BCL-MOS(V)-AT-PPO (Run3) &Train &$ 8.75 \pm 0.10 $ & $ 8.73 \pm 0.11 $ & $ 8.64 \pm 0.11 $ & $ 5.97 \pm 0.16 $	\\
& Eval &$ 4.64 \pm 0.16 $ & $ 4.63 \pm 0.16 $ & $ 4.49 \pm 0.16 $ & $ 4.14 \pm 0.16 $\\
\midrule
\rowcolor{Gray}BCL-MOS(R)-AT-PPO (Run1) &Train &$ 8.09 \pm 0.12 $ & $ 8.29 \pm 0.12 $ & $ 8.40 \pm 0.12 $ & $ 6.93 \pm 0.15 $ \\
\rowcolor{Gray}& Eval &$ 4.39 \pm 0.16 $ & $ 4.29 \pm 0.16 $ & $ 4.09 \pm 0.16 $ & $ 3.85 \pm 0.15 $ \\\cmidrule{1-1}   
BCL-MOS(R)-AT-PPO (Run2) &Train &$ 8.27 \pm 0.12 $ & $ 7.27 \pm 0.14 $ & $ 6.99 \pm 0.15 $ & $ 6.11 \pm 0.15 $\\
& Eval &$ 4.53 \pm 0.16 $ & $ 4.33 \pm 0.16 $ & $ 4.25 \pm 0.16 $ & $ 3.91 \pm 0.15 $ \\\cmidrule{1-1}   
BCL-MOS(R)-AT-PPO (Run3) &Train &$ 8.16 \pm 0.12 $ & $ 8.1 \pm 0.12 $ & $ 8.35 \pm 0.12 $ & $ 7.36 \pm 0.14 $\\
& Eval &$ 4.58 \pm 0.16 $ & $ 4.3 \pm 0.16 $ & $ 4.29 \pm 0.16 $ & $ 4.15 \pm 0.16 $	\\
\bottomrule
\end{tabular}
\end{sc}
\end{small}
\end{center}
\vskip -0.1in
\end{table}

\section{Hyperparameters for DQN models}
\label{app:thresholds_dqn}

\begin{table}[htbp!]
\caption{Thresholds $\overline{V}(\epsilon)$ for NCL/BCL-RADIAL-DQN}
\label{TH_RADIAL_DQN}
\vskip 0.15in
\begin{center}
\begin{small}
\begin{sc}
\begin{tabular}{lllll}
\toprule
  Criteria/Env.        & Pong & FW & BH & RR \\\midrule
  $\overline{V}_{\text{nominal}}(\epsilon)$   & 20   & 32      & 1200      & 35000     \\
\bottomrule
\end{tabular}
\end{sc}
\end{small}
\end{center}
\vskip -0.1in
\end{table}

\begin{table*}[!htbp]
\caption{BCL-AT specific hyperparameters}
\label{tab_bcl_params}
\vskip 0.15in
\begin{center}
\begin{small}
\begin{sc}
\begin{tabular}{lccc|ccc}
\toprule\toprule
Method/Env.            & \multicolumn{3}{c}{Pong} & \multicolumn{3}{c}{Freeway} \\\midrule
            & $K$   & $K_{\text{min}}$    & $\epsilon$ & $K$    & $K_{\text{min}}$     & $\epsilon$    \\\toprule
BCL-C-AT-DQN &   3  &    3         &  $\frac{30}{255}$    &  3  &3&$\frac{20}{255}$        \\\cmidrule{1-1}
BCL-MOS-AT-DQN &  3   &     1         &  $\frac{30}{255}$    &   3  &1&$\frac{20}{255}$     \\
\bottomrule\toprule 
Method/Env. &
 \multicolumn{3}{c}{BankHeist} & \multicolumn{3}{c}{RoadRunner} \\\midrule
&  $K$     & $K_{\text{min}}$    & $\epsilon$    & $K$     & $K_{\text{min}}$       &$\epsilon$ \\\toprule
BCL-C-AT-DQN &    5  &      5      &    $\frac{15}{255}$      &       3      &    3        &    $\frac{15}{255}$   \\\cmidrule{1-1}
BCL-MOS-AT-DQN    &5            &  1          &  $\frac{15}{255}$           &   3    &   1      & $\frac{15}{255}$\\
\cmidrule{1-1}
BCL-RADIAL+AT-DQN &  3   &     3         &  $\frac{15}{255}$    &   1  &1&$\frac{15}{255}$     \\\bottomrule
\end{tabular}
\end{sc}
\end{small}
\end{center}
\vskip -0.1in
\end{table*} 

\begin{table}[!htbp]
\caption{Thresholds $\overline{V}(\epsilon)$ for BCL-MOS-AT-DQN}
\label{mos_th}
\vskip 0.15in
\begin{center}
\begin{small}
\begin{sc}
\begin{tabular}{lllll}
\toprule
  Criteria/Env.        & Pong & FW & BH & RR \\\midrule
  $\overline{V}_{\text{nominal}}(\epsilon)$   & 20   & 30      & 1200      & 40000     \\
$\overline{V}_{\text{adv}}(\epsilon)$ & 20   & 25      & 1000      & 12000      \\
\bottomrule
\end{tabular}
\end{sc}
\end{small}
\end{center}
\vskip -0.1in
\end{table}

\section{Additional Experimental Results for DQN models}
\label{app:dqn_additional}
Below we show the detailed experimental results for all three independent runs for each experiment. We separate the results by environments for better visualization and comparison. That is, we have four tables, each for Pong (Table \ref{ap:Pong_BCL}), Freeway (Table \ref{ap:FW_BCL}), BankHeist (Table \ref{ap:BH_BCL}) and RoadRunner (Table \ref{ap:RR_BCL}). The discount factor used for evaluation is 1, with maximum episode length 10000, which is the same as in \citet{Oikarinen2021}. For benchmark models DQN (Vanilla) and SA-DQN (Convex) we use the released models from \citet{Zhang2020}, and for RADIAL-DQN we use the released models from \citet{Oikarinen2021}. For benchmark models AT-DQN and NCL-AT-DQN, we perform our own training as two restricted cases of the BCL algorithm. 

For AT-DQN we include all 9 runs for each environment, naming them as AT-DQN-$\epsilon$, where $\epsilon$ is the single value of the base curriculum taken as an input by the model. The choices of $\epsilon$ are the ones we use for evaluation. For each AT-DQN-$\epsilon$ we conduct three independent runs, and use the median of  those three runs as the representative result. The result reported in Table \ref{bcl_main_table}  for each environment is the best median result among those three $\epsilon$.

For NCL-AT-DQN and NCL-RADIAL-DQN we present the best result along the curriculum path for each run. For better comparison between different approaches, we set the target robustness level $\epsilon$ in curriculum for NCL-AT-DQN the same as in BCL experiments, that is, $\epsilon = 25/255$ for Pong, $\epsilon=20/255$ for Freeway, $\epsilon = 15/255$ for BankHeist and $\epsilon = 15/255$ for RoadRunner. %

\begin{table}[htbp!]
\caption{Pong environment. Average episode rewards $\pm$ standard error of the mean (SEM) over 20 episodes. The \colorbox{Gray}{gray rows} are the median of three runs (selected based on efficacy score $R_{\text{nominal}} + \frac{1}{3}\sum_\epsilon R_{\text{adv}}^{\epsilon}$). We report the lowest rewards among 30-step PGD, RI-FGSM, RI-FGSM (Multi) and RI-FGSM (Multi-T) attacks.}%
\label{ap:Pong_BCL}
\vskip 0.15in
\begin{center}
\begin{small}
\begin{sc}
\begin{tabular}{lllll}
\toprule\toprule
\multicolumn{5}{c}{\textbf{Pong}}\\
Model/Metric                & Nominal & \multicolumn{3}{c}{30-step PGD/RI-FGSM Attack}   \\\midrule
 $\epsilon$               & 0       & 10/255  & 20/255 & 25/255 \\\toprule
DQN (Vanilla)       &    ${21.0\pm 0.0}$  & $-21.0\pm 0.0$& $-21.0\pm 0.0$& $-21.0\pm 0.0$ \\
SA-DQN (Convex)        &    ${21.0\pm 0.0}$ & $-21.0\pm 0.0$& $-21.0\pm 0.0$&  $-21.0\pm 0.0$   \\
RADIAL-DQN     &${21.0\pm 0.0}$ &$-21.0\pm 0.0$&$-21.0\pm 0.0$&  $-21.0\pm 0.0$  \\\midrule
AT-DQN-10/255 (Run1) &$21.0 \pm 0.0 $&$ -21.0 \pm 0.0 $&$ -21.0 \pm 0.0 $&$-21.0\pm 0.0$\\
\rowcolor{Gray}AT-DQN-10/255 (Run2)  &${21.0 \pm 0.0 }$&$ -16.7 \pm 2.7 $& $ -21.0 \pm 0.0 $ & $ -20.9 \pm 0.1 $\\
\bottomrule
\end{tabular}
\end{sc}
\end{small}
\end{center}
\vskip -0.1in
\end{table}

\begin{table}[htbp!]
\caption{Detailed experiment results for Pong environment. The lowest reward under attacks are marked \colorbox{Gray}{gray}.}
\vskip 0.15in
\begin{center}
\begin{small}
\begin{sc}
    \begin{tabular}{lllll}
    \toprule\toprule
    \multicolumn{5}{c}{\textbf{Pong}}\\
Model                &   Attack    & $\epsilon$ = 10/255  &$\epsilon$= 20/255 & $\epsilon$ =25/255 \\\toprule
DQN (Vanilla) &30-step PGD &\ccolor$-21.0 \pm 0.0$ & \ccolor$-21.0 \pm 0.0$& \ccolor$ -21.0 \pm 0.0 $\\
 SA-DQN (Convex) &30-step PGD  &\ccolor$-21.0 \pm 0.0$ & \ccolor$-21.0 \pm 0.0$&\ccolor$ -21.0 \pm 0.0 $ \\
 RADIAL-DQN&30-step PGD  &\ccolor$-21.0 \pm 0.0$ & \ccolor$-21.0 \pm 0.0$& \ccolor$ -21.0 \pm 0.0 $\\\midrule
      AT-DQN-10/255 (Run1) &30-step PGD &\ccolor$-21.0 \pm 0.0$ & \ccolor$-21.0 \pm 0.0$&\ccolor$ -21.0 \pm 0.0 $\\\cmidrule{1-1}
 AT-DQN-10/255 (Run2) &30-step PGD  &\ccolor$ -16.7 \pm 2.7 $&\ccolor$ -21.0 \pm 0.0 $&$ -20.8 \pm 0.1 $\\
       & RI-FGSM &$ 20.1 \pm 0.0 $ & $ -20.8 \pm 0.1 $ &$ -20.8 \pm 0.1 $ \\
\bottomrule
    \end{tabular}
\end{sc}
\end{small}
\end{center}
\vskip -0.1in
\end{table}

\begin{table}[htbp!]
\caption{Detailed experiment results for Pong environment. The lowest reward under attacks are marked \colorbox{Gray}{gray}.}
\vskip 0.15in
\begin{center}
\begin{small}
\begin{sc}
\begin{tabular}{lllll}
    \toprule\toprule
    \multicolumn{5}{c}{\textbf{Pong}}\\
Model                &   Attack    & $\epsilon$ = 10/255  &$\epsilon$= 20/255 & $\epsilon$ =25/255 \\\toprule
      NCL-AT-DQN (Run1) &30-step PGD &\ccolor$-21.0 \pm 0.0$ & \ccolor$-21.0 \pm 0.0$&\ccolor$ -21.0 \pm 0.0 $\\\cmidrule{1-1}
NCL-AT-DQN (Run2)&30-step PGD  &$21.0 \pm 0.0 $&\ccolor$ -21.0 \pm 0.0 $&\ccolor$ -21.0 \pm 0.0 $\\
        & RI-FGSM  &$ 20.9 \pm 0.0 $ & $ -20.1 \pm 0.2 $ &$ -20.9 \pm 0.0 $ \\
        & RI-FGSM (Multi) &\ccolor$ 20.4 \pm 0.2 $ &\ccolor $ -21.0 \pm 0.0 $ & \ccolor$ -21.0 \pm 0.0 $ \\
        & RI-FGSM (Multi-T)  &$ 20.9 \pm 0.1 $ &\ccolor $ -21.0 \pm 0.0 $ &\ccolor$ -21.0 \pm 0.0 $ \\\cmidrule{1-1}
NCL-AT-DQN (Run3)&30-step PGD  &\ccolor$21.0 \pm 0.0 $&\ccolor$ -21.0 \pm 0.0 $&\ccolor$ -21.0 \pm 0.0 $\\
\bottomrule
    \end{tabular}
\end{sc}
\end{small}
\end{center}
\vskip -0.1in
\end{table}

\begin{table}[htbp!]
\caption{Detailed experiment results for Pong environment. The lowest reward under attacks are marked \colorbox{Gray}{gray}.}
\vskip 0.15in
\begin{center}
\begin{small}
\begin{sc}
   \begin{tabular}{lllll}
    \toprule\toprule
    \multicolumn{5}{c}{\textbf{Pong}}\\
Model                &   Attack    & $\epsilon$ = 10/255  &$\epsilon$= 20/255 & $\epsilon$ =25/255 \\\toprule
      NCL-RADIAL-DQN (Run1) & 30-step PGD &\ccolor$21.0 \pm 0.0$ & $12.9 \pm 2.0$ & $ 12.8 \pm 2.0 $\\
        & RI-FGSM&\ccolor$21.0 \pm 0.0$ &\ccolor $-21.0 \pm 0.0$ & \ccolor$ -21.0 \pm 0.0 $ \\
        & RI-FGSM (Multi) &\ccolor$ 21.0 \pm 0.0 $ & \ccolor$ -21.0 \pm 0.0 $ &\ccolor $ -21.0 \pm 0.0 $ \\
        & RI-FGSM (Multi-T) &\ccolor$21.0 \pm 0.0$ & \ccolor$-21.0 \pm 0.0$ & \ccolor$ -21.0 \pm 0.0 $\\\cmidrule{1-1}
NCL-RADIAL-DQN (Run2) &30-step PGD &\ccolor$-20.6 \pm 0.1$ &\ccolor $-21.0 \pm 0.0$ &\ccolor$ -21.0 \pm 0.0 $ \\
        & RI-FGSM&$-6.7 \pm 1.1$ &\ccolor $-21.0 \pm 0.0$ &\ccolor$ -21.0 \pm 0.0 $\\
\bottomrule
    \end{tabular}
\end{sc}
\end{small}
\end{center}
\vskip -0.1in
\end{table}

\begin{table}[htbp!]
\caption{Freeway environment. Average episode rewards $\pm$ standard error of the mean (SEM) over 20 episodes. The \colorbox{Gray}{gray rows} are the median of three runs (selected based on efficacy score $R_{\text{nominal}} + \frac{1}{3}\sum_\epsilon R_{\text{adv}}^{\epsilon}$). We report the lowest rewards among 30-step PGD, RI-FGSM, RI-FGSM (Multi) and RI-FGSM (Multi-T) attacks.} %
\label{ap:FW_BCL}
\vskip 0.15in
\begin{center}
\begin{small}
\begin{sc}
\begin{tabular}{lllll}
\toprule\toprule
\multicolumn{5}{c}{\textbf{Freeway}}\\
  Model/Metric              & Nominal & \multicolumn{3}{c}{30-step PGD/RI-FGSM Attack}   \\\midrule
 $\epsilon$               & 0      & 10/255  & 15/255 & 20/255 \\\toprule
DQN (Vanilla)        &   ${33.9 \pm 0.1} $&$ 0.0  \pm 0.0 $&$ 0.0  \pm 0.0 $&$ 0.0  \pm 0.0 $    \\
SA-DQN (Convex)       &$30.0 \pm 0.0 $&$ 19.3 \pm 0.4 $& $19.3 \pm 0.3 $&$ 20.0 \pm 0.3 $   \\
RADIAL-DQN      &$33.2 \pm 0.2 $&$ 17.1 \pm 0.3 $ & $ 13.4 \pm 0.2 $ & $ 7.9  \pm 0.3 $   \\\midrule
\rowcolor{Gray}AT-DQN-10/255 (Run1) &$32.4 \pm 0.2 $&$ 0.0 \pm 0.0 $&$ 0.0 \pm 0.0 $&$ 0.0 \pm 0.0$\\
AT-DQN-10/255 (Run2) &$33.3 \pm 0.2 $&$ 0.0 \pm 0.0 $&$ 0.0 \pm 0.0 $&$ 0.0 \pm 0.0$\\
\bottomrule
\end{tabular}
\end{sc}
\end{small}
\end{center}
\vskip -0.1in
\end{table}

\begin{table}[htbp!]
    \caption{Detailed experiment results for Freeway environment. The lowest reward under attacks are marked \colorbox{Gray}{gray}.}
\vskip 0.15in
\begin{center}
\begin{small}
\begin{sc}
    \begin{tabular}{lllll}
    \toprule\toprule
    \multicolumn{5}{c}{\textbf{Freeway}}\\
Model                &   Attack    & $\epsilon$ = 10/255  &$\epsilon$= 15/255 & $\epsilon$ =20/255 \\\toprule
DQN (Vanilla) &30-step PGD &\ccolor $  0.0 \pm 0.0 $&\ccolor$ 0.0 \pm 0.0 $&\ccolor$ 0.0 \pm 0.0$\\\cmidrule{1-1}
 SA-DQN (Convex) &30-step PGD  &\ccolor$ 19.3 \pm 0.4 $&\ccolor$ 19.3 \pm 0.3 $&\ccolor$ 20.0 \pm 0.3 $     \\
       & RI-FGSM &$ 21.2 \pm 0.4 $ & $ 21.5 \pm 0.4 $ & $ 21.9 \pm 0.3 $\\
       & RI-FGSM (Multi) &$ 21.4 \pm 0.3 $ & $ 21.3 \pm 0.3 $ & $ 20.7 \pm 0.5 $ \\
        & RI-FGSM (Multi-T)   &$ 21.4 \pm 0.3 $ & $ 21.1 \pm 0.3 $ & $ 21.4 \pm 0.3 $\\\cmidrule{1-1}
 RADIAL-DQN&30-step PGD  &$ 19.9 \pm 0.3 $&\ccolor$ 13.4 \pm 0.2 $&\ccolor$ 7.9  \pm 0.3 $   \\
\bottomrule
    \end{tabular}
\end{sc}
\end{small}
\end{center}
\vskip -0.1in
\end{table}

\begin{table}[htbp!]
\caption{Detailed experiment results for Freeway environment. The lowest reward under attacks are marked \colorbox{Gray}{gray}.}
\vskip 0.15in
\begin{center}
\begin{small}
\begin{sc}
    \begin{tabular}{lllll}
    \toprule\toprule
    \multicolumn{5}{c}{\textbf{Freeway}}\\
Model                &   Attack    & $\epsilon$ = 10/255  &$\epsilon$= 15/255 & $\epsilon$ =20/255 \\\toprule
      NCL-AT-DQN (Run1) &30-step PGD &$ {22.8 \pm 0.4 }$&$ 17.0 \pm 0.4 $&\ccolor$ 0.0 \pm 0.0 $\\
        & RI-FGSM  &$29.7 \pm 0.3$ & $29.9 \pm 0.4$ & $22.3 \pm 0.3$\\
        & RI-FGSM (Multi) &$ 23.6 \pm 0.5 $ & $ 11.3 \pm 0.4 $ &\ccolor $ 0.0 \pm 0.0 $ \\
        & RI-FGSM (Multi-T)  &\ccolor$22.0 \pm 0.5$ & \ccolor$9.6 \pm 0.4$ & \ccolor$0.0 \pm 0.0$\\\cmidrule{1-1}
NCL-AT-DQN (Run2) &30-step PGD  &\ccolor$ 26.2 \pm 0.2 $&\ccolor$ 17.9 \pm 0.3 $&\ccolor$ 3.9 \pm 0.2 $\\
       & RI-FGSM &$30.2 \pm 0.3$ & $30.4 \pm 0.4$ & $30.4 \pm 0.4$\\
\bottomrule
    \end{tabular}
\end{sc}
\end{small}
\end{center}
\vskip -0.1in
\end{table}

\begin{table}[htbp!]
\caption{Detailed experiment results for Freeway environment. The lowest reward under attacks are marked \colorbox{Gray}{gray}.}
\vskip 0.15in
\begin{center}
\begin{small}
\begin{sc}
    \begin{tabular}{lllll}
    \toprule\toprule
    \multicolumn{5}{c}{\textbf{Freeway}}\\
Model                &   Attack    & $\epsilon$ = 10/255  &$\epsilon$= 15/255 & $\epsilon$ =20/255 \\\toprule
      NCL-RADIAL-DQN (Run1) &30-step PGD &$25.4 \pm 0.4$ & $15.8 \pm 0.3$ &\ccolor $13.0 \pm 0.5$\\
        & RI-FGSM &$25.8 \pm 0.5$ & $20.6 \pm 0.4$ & $21.8 \pm 0.3$\\
        & RI-FGSM (Multi) &\ccolor$ 11.9 \pm 0.5 $ & \ccolor$ 0.2 \pm 0.1 $ & $ 21.5 \pm 0.4 $ \\
        & RI-FGSM (Multi-T) &$12.1 \pm 0.5$ & $0.3 \pm 0.1$ & $21.3 \pm 0.4$\\\cmidrule{1-1}
NCL-RADIAL-DQN (Run2) &30-step PGD &\ccolor$33.1 \pm 0.1$ & $23.3 \pm 0.5$ & $22.5 \pm 0.3$\\
       & RI-FGSM&$33.5 \pm 0.1$ & \ccolor$22.5 \pm 0.5$ &\ccolor $21.6 \pm 0.4$\\
\bottomrule
    \end{tabular}
\end{sc}
\end{small}
\end{center}
\vskip -0.1in
\end{table}

\begin{table}[htbp!]
\caption{BankHeist environment. Average episode rewards $\pm$ standard error of the mean (SEM) over 20 episodes. The \colorbox{Gray}{gray rows} are the median of three runs (selected based on efficacy score $R_{\text{nominal}} + \frac{1}{3}\sum_\epsilon R_{\text{adv}}^{\epsilon}$). We report the lowest rewards among 30-step PGD, RI-FGSM, RI-FGSM (Multi) and RI-FGSM (Multi-T) attacks.}%
\label{ap:BH_BCL}
\vskip 0.15in
\begin{center}
\begin{small}
\begin{sc}
\begin{tabular}{lllll}
\toprule\toprule
\multicolumn{5}{c}{\textbf{BankHeist}}\\
  Model/Metric              & Nominal & \multicolumn{3}{c}{30-step PGD/RI-FGSM Attack}   \\\midrule
 $\epsilon$               & 0         & 5/255   & 10/255 & 15/255\\\toprule
DQN (Vanilla)              & $1325.5  \pm 5.7$      & $0.0    \pm 0.0$      & $0.0    \pm 0.0$     & $0.0    \pm 0.0$      \\
SA-DQN (Convex) & $1237.5 \pm 1.7$   & $ 1126.0 \pm 32.0 $ & $ 63.0 \pm 3.5 $ &  $ 16.0 \pm 1.6 $\\
RADIAL-DQN      & ${1349.5  \pm 1.7}$    & $ 581.5 \pm 16.7 $    & $0.0    \pm 0.0$     &    $0.0    \pm 0.0$   \\\midrule
\rowcolor{Gray}AT-DQN-5/255 (Run1) &$1200.0 \pm 12.1$&$95.5\pm5.9$&$0.0 \pm 0.0$ & $0.0  \pm 0.0$\\
AT-DQN-5/255 (Run2) &$1217.0 \pm 10.1 $  &$407.5\pm 30.6$ &  $4.5 \pm 1.1$ & $1.0\pm 0.7$\\
\bottomrule
\end{tabular}
\end{sc}
\end{small}
\end{center}
\vskip -0.1in
\end{table}

\begin{table}[htbp!]
    \caption{Detailed experiment results for BankHeist environment. The lowest reward under attacks are marked \colorbox{Gray}{gray}.}
\vskip 0.15in
\begin{center}
\begin{small}
\begin{sc}
    \begin{tabular}{lllll}
    \toprule\toprule
    \multicolumn{5}{c}{\textbf{BankHeist}}\\
Model                &   Attack    & $\epsilon$ = 5/255  &$\epsilon$= 10/255 & $\epsilon$ =15/255 \\\toprule
DQN (Vanilla) &30-step PGD &\ccolor$0.0    \pm 0.0$      & \ccolor$0.0    \pm 0.0$     &\ccolor $0.0    \pm 0.0$      \\\cmidrule{1-1}
 SA-DQN (Convex) &30-step PGD  &$1155.5\pm6.4$ & $114.0\pm3.8$ &$ 24.5 \pm 3.1 $\\
       & RI-FGSM &$ 1153.5 \pm 29.5 $ & $ 556.0 \pm 25.1 $ & $ 137.5 \pm 8.3 $\\
       & RI-FGSM (Multi) &\ccolor$ 1126.0 \pm 32.0 $ & \ccolor$ 63.0 \pm 3.5 $ & $ 20.0 \pm 1.4 $ \\
        & RI-FGSM (Multi-T)  & $ 1128.0 \pm 31.3 $ &  $ 75.5 \pm 4.4 $ & \ccolor $ 16.0 \pm 1.6 $\\\cmidrule{1-1}
 RADIAL-DQN&30-step PGD  &$761.0  \pm 24.5$    &\ccolor $0.0    \pm 0.0$     &   \ccolor $0.0    \pm 0.0$\\
\bottomrule
    \end{tabular}
\end{sc}
\end{small}
\end{center}
\vskip -0.1in
\end{table}

\begin{table}[htbp!]
    \caption{Detailed experiment results for BankHeist environment. The lowest reward under attacks are marked \colorbox{Gray}{gray}.}
\vskip 0.15in
\begin{center}
\begin{small}
\begin{sc}
    \begin{tabular}{lllll}
    \toprule\toprule
    \multicolumn{5}{c}{\textbf{BankHeist}}\\
Model                &   Attack    & $\epsilon$ = 5/255  &$\epsilon$= 10/255 & $\epsilon$ =15/255 \\\toprule
      NCL-AT-DQN (Run1) &30-step PGD &\ccolor$ 245.0 \pm 23.7 $&\ccolor$ 1.0 \pm 0.7$ &\ccolor$ 0.0  \pm 0.0 $\\
        & RI-FGSM  &$ 1080.0 \pm 16.6 $ & $ 34.0 \pm 7.8 $ & $ 22.5 \pm 2.0 $\\
        & RI-FGSM (Multi) &$ 1003.5 \pm 22.7 $ & $ 38.5 \pm 4.7 $ & $ 15.5 \pm 3.0 $ \\
        & RI-FGSM (Multi-T)  &$ 1025.5 \pm 25.5 $ & $ 28.5 \pm 6.9 $ & $ 28.0 \pm 3.4 $\\\cmidrule{1-1}
NCL-AT-DQN (Run2) &30-step PGD  &\ccolor$ 3.0   \pm 1.0  $&\ccolor$ 0.0  \pm 0.0 $&\ccolor$ 0.0 \pm 0.0$\\
       & RI-FGSM &$ 215.0 \pm 42.4 $ & $ 29.0 \pm 3.4 $ & $ 7.0 \pm 1.7 $\\
\bottomrule
    \end{tabular}
\end{sc}
\end{small}
\end{center}
\vskip -0.1in
\end{table}

\begin{table}[htbp!]
    \caption{Detailed experiment results for BankHeist environment. The lowest reward under attacks are marked \colorbox{Gray}{gray}.}
\vskip 0.15in
\begin{center}
\begin{small}
\begin{sc}
    \begin{tabular}{lllll}
    \toprule\toprule
    \multicolumn{5}{c}{\textbf{BankHeist}}\\
Model                &   Attack    & $\epsilon$ = 5/255  &$\epsilon$= 10/255 & $\epsilon$ =15/255 \\\toprule
      NCL-RADIAL-DQN (Run1) & 30-step PGD &$ 1272.0 \pm 2.9 $ & $ 1264.0 \pm 4.1 $ & \ccolor$ 45.5 \pm 2.1 $\\
        & RI-FGSM &$ 1266.0 \pm 7.0 $ & $ 1253.5 \pm 7.1 $ & $ 968.0 \pm 16.6 $\\
        & RI-FGSM (Multi) &\ccolor$ 1265.0 \pm 5.6 $ & \ccolor$ 1243.0 \pm 7.5 $ & $ 810.5 \pm 14.4 $ \\
        & RI-FGSM (Multi-T) &\ccolor$ 1265.0 \pm 5.6 $ & \ccolor$ 1243.0 \pm 7.5 $ & $ 822.0 \pm 18.0 $\\\cmidrule{1-1}
NCL-RADIAL-DQN (Run2)  &30-step PGD &$1242.5 \pm 3.5$ & $89.5 \pm 9.1$ &\ccolor $9.0 \pm 1.9$\\
& RI-FGSM &$1226.5 \pm 2.9$ & $69.0 \pm 10.8$ & $11.5 \pm 1.1$\\
\bottomrule
    \end{tabular}
\end{sc}
\end{small}
\end{center}
\vskip -0.1in
\end{table}

\begin{table}[htbp!]
\caption{RoadRunner environment. Average episode rewards $\pm$ standard error of the mean (SEM) over 20 episodes. The \colorbox{Gray}{gray rows} are the median of three runs (selected based on efficacy score $R_{\text{nominal}} + \frac{1}{3}\sum_\epsilon R_{\text{adv}}^{\epsilon}$). We report the lowest rewards among 30-step PGD, RI-FGSM, RI-FGSM (Multi) and RI-FGSM (Multi-T) attacks.}%
\label{ap:RR_BCL}
\vskip 0.15in
\begin{center}
\begin{small}
\begin{sc}
\begin{tabular}{lllll}
\toprule\toprule
\multicolumn{5}{c}{\textbf{RoadRunner}}\\
   Model/Metric             & Nominal & \multicolumn{3}{c}{30-step PGD/RI-FGSM Attack}  \\\midrule
 $\epsilon$               & 0         & 5/255 & 10/255 & 15/255 \\\toprule
DQN (Vanilla)            & $43390   \pm 973$        & $0     \pm 0$     & $0     \pm 0$      & $0     \pm 0$      \\
SA-DQN (Convex)       & $45870   \pm 1380$    & $985   \pm 207$   & $0     \pm 0$      & $0     \pm 0$      \\
RADIAL-DQN      & $44595   \pm 1165$    & $7195  \pm 929$  & $495    \pm 116$    & $ 0 \pm 0 $     \\
\midrule
AT-DQN-5/255 (Run1) &$ 44065 \pm 1896 $ & $ 13150 \pm 2116 $ & $ 0 \pm 0 $ & $ 0 \pm 0 $\\
\rowcolor{Gray}AT-DQN-5/255 (Run2)&$ 39890 \pm 2092 $ & $ 20160 \pm 1973 $ & $ 0 \pm 0 $ & $ 0 \pm 0 $\\
\bottomrule
\end{tabular}
\end{sc}
\end{small}
\end{center}
\vskip -0.1in
\end{table}

\begin{table}[htbp!]
\caption{Detailed experiment results for RoadRunner environment. The lowest reward under attacks are marked \colorbox{Gray}{gray}.}
\vskip 0.15in
\begin{center}
\begin{small}
\begin{sc}
    \begin{tabular}{lllll}
    \toprule\toprule
    \multicolumn{5}{c}{\textbf{RoadRunner}}\\
Model                &   Attack    & $\epsilon$ = 5/255  &$\epsilon$= 10/255 & $\epsilon$ =15/255 \\\toprule
DQN (Vanilla) &30-step PGD &\ccolor$0   \pm 0$      &\ccolor $0   \pm 0$     &\ccolor $0   \pm 0$ \\\cmidrule{1-1}
 SA-DQN (Convex) &30-step PGD  &\ccolor$985   \pm 207$   & \ccolor$0     \pm 0$      & \ccolor$0     \pm 0$\\
       & RI-FGSM &$ 9740 \pm 677 $ & $ 5615 \pm 645 $ & $ 3575 \pm 490 $ \\
       & RI-FGSM (Multi) &$ 6170 \pm 759 $ & $ 1045 \pm 307 $ & $ 105 \pm 56 $ \\
        & RI-FGSM (Multi-T)  &$ 6870 \pm 833 $ & $ 1220 \pm 318 $ & $ 55 \pm 25 $\\\cmidrule{1-1}
 RADIAL-DQN&30-step PGD  &\ccolor$7195  \pm 929$  &\ccolor $495    \pm 116$    &\ccolor $ 0 \pm 0 $\\
\bottomrule
    \end{tabular}
\end{sc}
\end{small}
\end{center}
\vskip -0.1in
\end{table}

\begin{table}[htbp!]
    \caption{Detailed experiment results for RoadRunner environment. The lowest reward under attacks are marked \colorbox{Gray}{gray}.}
\vskip 0.15in
\begin{center}
\begin{small}
\begin{sc}
    \begin{tabular}{lllll}
    \toprule\toprule
    \multicolumn{5}{c}{\textbf{RoadRunner}}\\
Model                &   Attack    & $\epsilon$ = 5/255  &$\epsilon$= 10/255 & $\epsilon$ =15/255 \\\toprule
      NCL-AT-DQN (Run1) &30-step PGD &\ccolor$ 40235 \pm 2261 $ & \ccolor$ 1100 \pm 234 $ &\ccolor $ 0 \pm 0 $\\
        & RI-FGSM  &$ 45595 \pm 1781 $ & $ 13750 \pm 1118 $ & $ 5235 \pm 319 $\\
        & RI-FGSM (Multi) &$ 45645 \pm 1113 $ & $ 10925 \pm 1098 $ & $ 4030 \pm 328 $\\
        & RI-FGSM (Multi-T)  &$ 44035 \pm 1818 $ & $ 5860 \pm 812 $ & $ 985 \pm 228 $\\\cmidrule{1-1}
NCL-AT-DQN (Run2) &30-step PGD  &\ccolor$ 39045 \pm 2382 $ & \ccolor$ 15 \pm 8 $ & \ccolor$ 5 \pm 5 $\\
       & RI-FGSM &$ 41365 \pm 1850 $ & $ 14235 \pm 1387 $ & $ 4480 \pm 408 $\\
\bottomrule
    \end{tabular}
\end{sc}
\end{small}
\end{center}
\vskip -0.1in
\end{table}

\begin{table}[htbp!]
\caption{Detailed experiment results for RoadRunner environment. The lowest reward under attacks are marked \colorbox{Gray}{gray}.}
\vskip 0.15in
\begin{center}
\begin{small}

\begin{threeparttable}
  \begin{sc}  
    \begin{tabular}{lllll}
    \toprule\toprule
    \multicolumn{5}{c}{\textbf{RoadRunner}}\\
Model                &   Attack    & $\epsilon$ = 5/255  &$\epsilon$= 10/255 & $\epsilon$ =15/255 \\\toprule
NCL-RADIAL-DQN (Run3)\tnote{*}&30-step PGD &$ 41230 \pm 1920 $ & $ 41230 \pm 1920 $ & $ 41270 \pm 1926 $\\
        & RI-FGSM &\ccolor$ 40885 \pm 1921 $ & $ 37275 \pm 2212 $ & $ 9935 \pm 783 $\\
        & RI-FGSM (Multi) &\ccolor$ 40885 \pm 1921 $ & $ 18845 \pm 966 $ & $ 6325 \pm 446 $ \\
        & RI-FGSM (Multi-T) &\ccolor$ 40885 \pm 1921 $ & \ccolor$ 17050 \pm 1092 $ & \ccolor$ 6100 \pm 428 $\\\midrule
     BCL-RADIAL-DQN (Run1) &30-step PGD & $ 41045 \pm 1289 $ & $ 41045 \pm 1289 $ & $ 41045 \pm 1289 $ \\
     & RI-FGSM & $ 41045 \pm 1289 $ & \ccolor$ 37865 \pm 1082 $ & $ 12120 \pm 639 $\\
\bottomrule
    \end{tabular}
    \end{sc}
    \begin{tablenotes}\footnotesize
\item[*]NCL-RADIAL-DQN (Run1) \& (Run2) are the same as BCL-RADIAL-DQN (Run1) \& (Run2).
\end{tablenotes}
\end{threeparttable}
\end{small}
\end{center}
\vskip -0.1in
\end{table}

\section{Walltime for DQN models}
\label{app:dqn_walltime}
The training time for each run takes around  12 hours on a single GeForce RTX 2080Ti GPU. Below in Table \ref{ap:walltime_DQN} we present the number of runs conducted for each experiment. 

\begin{table}[htbp!]
\caption{Number of runs conducted for each experiment.}%
\label{ap:walltime_DQN}
\vskip 0.15in
\begin{center}
\begin{small}
\begin{sc}
\begin{tabular}{llll}
\toprule\toprule
\multicolumn{4}{c}{Pong}               \\
                  & Run1 & Run2 & Run3 \\\midrule
AT-DQN-$\epsilon$        & 1    & 1    & 1    \\
NCL-AT-DQN        & 25   & 25   & 25   \\
NCL-RADIAL-DQN    & 10   & 4    & 4    \\
BCL-C-AT-DQN      & 66   & 66   & 66   \\
BCL-MOS-AT-DQN    & 20   & 18   & 20   \\
BCL-RADIAL-DQN    & 8    & 21   & 29   \\
\bottomrule\toprule
\multicolumn{4}{c}{Freeway}            \\
                  & Run1 & Run2 & Run3 \\\midrule
AT-DQN-$\epsilon$        & 1    & 1    & 1    \\
NCL-AT-DQN        & 20   & 20   & 20   \\
NCL-RADIAL-DQN    & 6    & 12   & 7    \\
BCL-C-AT-DQN      & 51   & 51   & 51   \\
BCL-MOS-AT-DQN    & 25   & 26   & 21   \\
BCL-RADIAL-DQN    & 20   & 18   & 23   \\
                  \bottomrule\toprule
\multicolumn{4}{c}{BankHeist}          \\
                  & Run1 & Run2 & Run3 \\\midrule
AT-DQN-$\epsilon$        & 1    & 1    & 1    \\
NCL-AT-DQN        & 15   & 15   & 15   \\
NCL-RADIAL-DQN    & 10   & 9    & 9    \\
BCL-C-AT-DQN      & 60   & 60   & 60   \\
BCL-MOS-AT-DQN    & 38   & 43   & 44   \\
BCL-RADIAL-DQN    & 18   & 15   & 13   \\
BCL-RADIAL+AT-DQN & 24   & N/A  & N/A \\
                  \bottomrule\toprule
\multicolumn{4}{c}{RoadRunner}         \\
                  & Run1 & Run2 & Run3 \\\midrule
AT-DQN-$\epsilon$        & 1    & 1    & 1    \\
NCL-AT-DQN        & 15   & 15   & 15   \\
NCL-RADIAL-DQN    & 13   & 13   & 8    \\
BCL-C-AT-DQN      & 36   & 36   & 36   \\
BCL-MOS-AT-DQN    & 22   & 21   & 18   \\
BCL-RADIAL-DQN    & 17   & 17   & 17   \\
BCL-RADIAL+AT-DQN & 20   & N/A  & N/A \\
\bottomrule
\end{tabular}

\end{sc}
\end{small}
\end{center}
\vskip -0.1in
\end{table}
\end{document}